# Communication-efficient Distributed Estimation and Inference for Transelliptical Graphical Models[*]

Pan Xu[†] and Lu Tian[‡] and Quanquan Gu[§]

**Abstract**

We propose communication-efficient distributed estimation and inference methods for the transelliptical graphical model, a semiparametric extension of the elliptical distribution in the high dimensional regime. In detail, the proposed method distributes the $d$-dimensional data of size $N$ generated from a transelliptical graphical model into $m$ worker machines, and estimates the latent precision matrix on each worker machine based on the data of size $n = N/m$. It then debiases the local estimators on the worker machines and sends them back to the master machine. Finally, on the master machine, it aggregates the debiased local estimators by averaging and hard thresholding. We show that the aggregated estimator attains the same statistical rate as the centralized estimator based on all the data, provided that the number of machines satisfies $m \lesssim \min\{N \log d/d, \sqrt{N/(s^2 \log d)}\}$, where $s$ is the maximum number of nonzero entries in each column of the latent precision matrix. It is worth noting that our algorithm and theory can be directly applied to Gaussian graphical models, Gaussian copula graphical models and elliptical graphical models, since they are all special cases of transelliptical graphical models. Thorough experiments on synthetic data back up our theory.

## 1 Introduction

In the past two decades, a wide range of researches has been using graphical models (Lauritzen, 1996) to explore the dependence structure of multivariate distributions. For instance, in Gaussian graphical models (Yuan and Lin, 2007), a $d$-dimensional random vector $\boldsymbol{X} = (X_1, \ldots, X_d)^\top \in \mathbb{R}^d$ follows a multivariate normal distribution $N(\boldsymbol{0}, \boldsymbol{\Sigma}^*)$, where the precision matrix (inverse of the covariance matrix) $\boldsymbol{\Theta}^* = \boldsymbol{\Sigma}^{*-1}$ encodes the conditional independence relationship of the marginal random variables, i.e., $X_j$ and $X_k$ are independent conditioned on the rest of the marginal random variables if and only if $\Theta^*_{jk} = 0$. Therefore, it is of great interest to estimate the precision matrix. A large body of literature has studied the estimation problem for precision matrix $\boldsymbol{\Theta}^*$ in the high dimensional setting, where the number of parameters is much larger than the sample size, i.e., $d \gg n$ (Meinshausen and Bühlmann, 2006; Yuan and Lin, 2007; Banerjee et al., 2008; Friedman et al., 2008; Rothman et al., 2008; Yuan, 2010; Liu et al., 2010; Ravikumar et al., 2011; Cai et al., 2011; Loh and Wainwright, 2013; Zhao and Liu, 2013; Sun and Zhang, 2013; Wang et al., 2016; Cai et al., 2016). Despite the popularity and the appealing theoretical properties of Gaussian graphical models, the normality assumption is too restrictive. To relax the normality assumption, Liu et al. (2009, 2012a); Xue and Zou (2012) extended the Gaussian graphical model to a larger family of distributions namely Gaussian copula graphical models. Furthermore, Liu et al. (2012b) proposed the transelliptical distribution, which is a generalization of the elliptical distribution (Fang et al., 1990), just like how the Gaussian copula distribution family generalizes the Gaussian distribution. Specifically, if a $d$-dimensional random vector $\boldsymbol{X} = (X_1, \ldots, X_d)^\top$ follows a transelliptical distribution, then we denote it by $\boldsymbol{X} \sim TE_d(\boldsymbol{\Sigma}^*, \xi; f_1, \ldots, f_d)$, where $\boldsymbol{\Theta}^* = (\boldsymbol{\Sigma}^*)^{-1}$ is the latent precision matrix, $f_1, \ldots, f_d$ are monotone univariate functions and $\xi$ is a nonnegative random variable. Liu et al. (2012b) showed that the latent precision matrix $\boldsymbol{\Theta}^*$ is able to encode the conditional independence of the marginal random variables $X_1, \ldots, X_d$ in transelliptical graphical models. This motivates the estimation of the latent precision matrix for transelliptical graphical models in practice. Note that transelliptical graphical models include Gaussian graphical models, Gaussian copula graphical models and elliptical graphical models (Liu et al., 2009, 2012a; Xue and Zou, 2012; Wegkamp et al., 2016) as special cases.

---

[†]Department of Systems and Information Engineering, University of Virginia, Charlottesville, VA 22904, USA; e-mail:px3ds@virginia.edu

[‡]Department of Systems and Information Engineering, University of Virginia, Charlottesville, VA 22904, USA; e-mail:lt2eu@virginia.edu

[§]Department of Systems and Information Engineering, Department of Computer Science, University of Virginia, Charlottesville, VA 22904, USA; e-mail:qg5w@virginia.edu

Besides the aforementioned estimation problem for high dimensional graphical models, quite a few asymptotic inference methods including hypothesis testing and confidence intervals construction (Jankova and van de Geer, 2013; Liu et al., 2013; Van de Geer et al., 2014; Ning and Liu, 2014; Yang et al., 2014; Janková and van de Geer, 2015; Ren et al., 2015; Gu et al., 2015; Barber and Kolar, 2015; Neykov et al., 2015; Chen et al., 2016) have also been proposed for high dimensional graphical models. For instance, Jankova and van de Geer (2013); Janková and van de Geer (2015) proposed inference methods for low-dimensional parameters of sparse precision matrices in Gaussian graphical models based on debiased estimators (Javanmard and Montanari, 2013; Van de Geer et al., 2014). For semiparametric graphical models, Gu et al. (2015) and Barber and Kolar (2015) studied the inference for nonparanormal and transelliptical graphical models respectively. Based on the general framework for hypothesis tests and confidence regions proposed in Ning and Liu (2014), Yang et al. (2014) studied a class of semiparametric exponential family graphical models for high dimensional mixed data, and Neykov et al. (2015) studied a unified theory of statistical inference for high dimensional estimating equations, which also investigated the transelliptical graphical model as an example.

Nevertheless, all the above methods for point estimation and asymptotic inference of the (latent) precision matrices are based on a single machine. In many big data applications, however, the data are often collected and stored independently or they are too large to be processed on one machine. Issues arise in aggregating large datasets due to the ever increasing algorithmic difficulties. To address this, various distributed algorithms (Mcdonald et al., 2009; Zinkevich et al., 2010; Boyd et al., 2011; Zhang et al., 2012; Liu and Ihler, 2012; Rosenblatt and Nadler, 2014; Chen and Xie, 2014; Liu and Ihler, 2014; Lee et al., 2015; Battey et al., 2015; Arroyo and Hou, 2016; Tian and Gu, 2016) have been proposed in the machine learning, optimization and statistics communities, which divide the data into several machines, perform the estimation and inference locally, and then aggregate the local estimators or test statistics in a principled way. The main challenge in distributed learning is to keep the communication between machines efficient while ensuring a comparable statistical performance as the centralized methods.

In this paper, we propose a communication-efficient distributed method for latent precision matrix estimation and statistical inference for transelliptical graphical models. In our proposed algorithm, each "worker" machine computes a local unbiased estimator for the latent precision matrix and sends it to the "master" machine. After receiving all local estimators, the master machine then averages them to form an aggregated estimator. We also propose a distributed hypothesis test procedure to test whether a specific edge exists in the latent precision matrix. At the core of our algorithm is the unbiased estimator for transelliptical graphical models, and a refined analysis of Hoeffding decomposition (Hoeffding, 1948) for high dimensional U-statistics. In the proposed algorithm, each worker machine only needs to send a $d \times d$ matrix to the central master machine once. Thus our algorithm is extremely communication-efficient by requiring only one round of communication between the worker machines and the master machine. Moreover, our estimator has computational advantage over the centralized estimator. Since the computational cost of the latent correlation matrix based on rank-based test statistic (Liu et al., 2012b) grows quadratically as the sample size goes up, the distributed algorithm greatly reduces the computational overhead.

For the latent precision matrix estimation, we prove that the proposed distributed algorithm attains $O_p\big(\sqrt{\log d/N} + \sqrt{d/(N(n-1))} + sm\log d/N\big)$ convergence rate in terms of $\ell_{\infty,\infty}$ norm, where $N$ is the total sample size, $m$ is the number of machines, $d$ is the dimensionality, $n = N/m$, and $s = \max_{1\leq j\leq d}\sum_{k=1}^d \mathbb{1}(\Theta_{jk}^* \neq 0)$ is the row sparsity of the true precision matrix. This addresses an important question on the choice of machine numbers, in order to minimize the information loss due to the data parallelism. It can be shown that when the number of machines $m$ satisfies $m \lesssim \min\{N\log d/d, \sqrt{N/(s^2\log d)}\}$, our distributed algorithm achieves the same statistical rate as the centralized estimator (Cai et al., 2011; Ravikumar et al., 2011), which obtains $O(\sqrt{\log d/N})$ convergence rate in terms of $\ell_{\infty,\infty}$ norm. In addition, our estimator achieves model selection consistency under the same condition as the centralized method (Cai et al., 2011) that $\min_{j,k}|\Theta_{jk}^*| \gtrsim \sqrt{\log d/N}$. Furthermore, for each element in our distributed estimator of precision matrix, we establish its asymptotical normality when $n \gtrsim (s\log d)^2$. This scaling matches those of the centralized asymptotic inference methods for Gaussian (Jankova and van de Geer, 2013; Liu et al., 2013), nonparanormal (Gu et al., 2015) and transelliptical graphical models (Barber and Kolar, 2015), and has been shown to be optimal in Ren et al. (2015).

## 1.1 Organization of the Paper

The remainder of this paper is organized as follows: we briefly review the related work in Section 2 and introduce the transelliptical graphical model in Section 3. Then we present our distributed precision matrix estimation method, and distributed hypothesis test procedure and its asymptotic properties in Section 4. We provide our main theory in Section 5, and the proofs of the main theory are presented in Section 6. We also discuss the application of our algorithm to the Gaussian graphical models and provide the corresponding theory in Section 7. Section 8 shows the experimental results of both estimation and inference on synthetic data. Section 9 concludes this paper.

## 1.2 Notation

We summarize here the notations to be used throughout this paper. Let $\mathbf{A} = [A_{ij}] \in \mathbb{R}^{d \times d}$ be a $d \times d$ matrix and $\mathbf{x} = [x_1, \ldots, x_d]^\top \in \mathbb{R}^d$ be a $d$-dimensional vector. For $0 < q < \infty$, we define the $\ell_0$, $\ell_q$ and $\ell_\infty$ vector norms as $\|\mathbf{x}\|_0 = \sum_{i=1}^d \mathbb{1}(x_i \neq 0)$, $\|\mathbf{x}\|_q = \left(\sum_{i=1}^d |x_i|^q\right)^{1/q}$, and $\|\mathbf{x}\|_\infty = \max_{1 \leq i \leq d} |x_i|$, where $\mathbb{1}(\cdot)$ represents the indicator function. For a matrix $\mathbf{A} \in \mathbb{R}^{d \times d}$, we use the following notations for the matrix $\ell_q$, $\ell_{\infty,\infty}$ and Frobenius norms: $\|\mathbf{A}\|_q = \max_{\|\mathbf{x}\|_q = 1} \|\mathbf{A}\mathbf{x}\|_q$, $\|\mathbf{A}\|_{\infty,\infty} = \max_{1 \leq i,j \leq d} |A_{ij}|$, and $\|\mathbf{A}\|_F = \left(\sum_{ij} |A_{ij}|^2\right)^{1/2}$. We use $\mathbf{A}_{j*}$ and $\mathbf{A}_{*k}$ to denote the $j$-th row and the $k$-th column of matrix $\mathbf{A}$. For an index set $S \in \{1, 2, \ldots, d\} \times \{1, 2, \ldots, d\}$, $\mathbf{A}_S$ denotes the matrix whose support set is restricted on set $S$, i.e., $[A_S]_{jk} = A_{jk}$ for $(j, k) \in S$ and $[A_S]_{jk} = 0$ for $(j, k) \notin S$. For a symmetric matrix $\mathbf{A}$, we use $\mathbf{A} \succ \mathbf{0}$ to denote that $\mathbf{A}$ is positive definite. For a sequence of random variables $X_n$, we use $X_n \xrightarrow{p} X$ to represent that $X_n$ converges in probability to $X$, and $X_n \rightsquigarrow X$ for convergence in distribution to $X$. For sequences $f_n, g_n$, we write $f_n = O(g_n)$ if $|f_n| \leq C|g_n|$ for some $C > 0$ independent of $n$ and all $n > C$, and $f_n = o(g_n)$ if $\lim_{n \to \infty} f_n/g_n = 0$. We also make use of the notation $f_n \lesssim g_n$ ($f_n \gtrsim g_n$) if $f_n$ is less than (larger than) $g_n$ up to a constant. Let $\Phi(\cdot)$ denote the cumulative distribution function (CDF) of standard normal distribution, and $\Phi^{-1}(\cdot)$ denote its inverse function.

## 2 Related Work

There are several lines of research on distributed algorithms in the machine learning, statistics and optimization communities. Our study focuses on the data distribution, where data are distributed across different machines. An important issue that occurs in all distributed learning problems is to reduce the communication cost while keeping a good performance. To address this problem, Mcdonald et al. (2009) proposed the averaging approach: each machine generates a local estimator and sends it to the "master" machine where all local estimators are averaged to form an aggregated estimator. The non-asymptotic analysis of the aggregated estimator shows that averaging only reduces the variance of the estimator, but not the bias. Zinkevich et al. (2010) studied a variant of the averaging approach where each machine calculates a local estimator with stochastic gradient descent on a random subset of the dataset. They showed that their estimator converges to the centralized estimator. Zhang et al. (2012) investigated distributed empirical risk minimization (ERM), and showed that the mean squared error (MSE) of the averaged ERM is in the order of $O(N^{-1} + (m/N)^2)$, where $m$ is the number of machines and $N$ is the total sample size. Hence if we set $m \lesssim \sqrt{N}$ when $N$ grows, the averaged ERM will have comparable error bounds with the centralized ERM. Moreover, Zhang et al. (2013) studied distributed kernel ridge regression method using the divide and conquer strategy and a theoretical guarantee is provided about the proper choice of the number of worker machines. Balcan et al. (2012) studied the problem of PAC-learning from distributed data and analyzed the communication complexity in terms of different complexity measures of the hypothesis space.

The distributed estimators mentioned above mainly focus on estimation in the classical regime, i.e., the dimension of data remains fixed while sample size grows. In the high dimensional regime, existing studies have shown that averaging is ineffective. For example, Rosenblatt and Nadler (2014) studied the optimality of averaged ERM when $d, n \to \infty$, $d/n \to \mu \in (0, 1)$, where $n = N/m$ is the sample size on each machine. They showed the averaged ERM is suboptimal. Furthermore, the regularization is a widely used technique in high dimensional estimations, while it usually involves bias to the estimator. For example, the Lasso estimator (Tibshirani, 1996) is biased due to the $\ell_1$-norm penalty (Fan and Li, 2001). Since averaging only reduces variances, not bias, the performance of the averaged estimator is not better than the local estimator due to the aggregation of bias when averaging. To address this problem, Lee et al. (2015) proposed a distributed sparse regression method, which exploits the debiased estimators proposed in Javanmard and Montanari (2013); Van de Geer et al. (2014) for distributed sparse regression. Similar distributed regression methods are proposed by Battey et al. (2015) for both distributed statistical estimation and hypothesis testing. For distributed classification, Tian and Gu (2016) proposed a distributed sparse linear discriminant analysis based on Cai et al. (2011) and a debiased estimator for Dantzig selector type estimator.

For distributed graphical model estimation, Hsieh et al. (2012) proposed a divide-and-conquer procedure for precision matrix estimation, where they divided the estimation problem into subproblems and then used the solutions of the sub-problems to approximate the solution for the original problem. However, there is no statistical guarantee for their distributed algorithm. Meng et al. (2014) studied the precision matrix estimation where the dimensions were distributed and the structure of the graphical model was assumed to be known. Our work is orthogonal to theirs, because we consider distributing the samples instead of distributing the variables. We would like to point out that Arroyo and Hou (2016) proposed a distributed estimation method for the Gaussian graphical model, which is a special case of the transelliptical graphical model we studied in this paper. Our algorithm and theoretical analysis are more general than theirs. Moreover, we do not require the stringent irrepresentable condition (Ravikumar et al., 2011; Jankova and van de Geer, 2013) as they do.

## 3 Transelliptical Graphical Models

In this section, we briefly review the semiparametric elliptical graphical model, namely transelliptical graphical model, which was originally proposed in Liu et al. (2012b), and its centralized estimation method. We begin with the definition of elliptical distribution (Fang et al., 1990).

**Definition 3.1** (Elliptical distribution). Let $\boldsymbol{\mu} \in \mathbb{R}^d$ and $\boldsymbol{\Sigma}^* \in \mathbb{R}^{d \times d}$ with rank($\boldsymbol{\Sigma}^*$) = $q \leq d$. A random vector $\boldsymbol{X} \in \mathbb{R}^d$ follows an elliptical distribution, denoted by $EC_d(\boldsymbol{\mu}, \boldsymbol{\Sigma}^*, \xi)$, if it can be represented as $\boldsymbol{X} = \boldsymbol{\mu} + \xi \mathbf{A} \mathbf{U}$, where $\mathbf{A}$ is a deterministic matrix satisfying $\mathbf{A}^\top \mathbf{A} = \boldsymbol{\Sigma}^*$, $\mathbf{U}$ is a random vector uniformly distributed on the unit sphere in $\mathbb{R}^q$, and $\xi$ is a random variable independent of $\mathbf{U}$.

The elliptical distribution is a generalization of the multivariate normal distribution. Liu et al. (2009) proposed the nonparanormal distribution, which is a nonparametric extension of the normal distribution. The following is the definition:

**Definition 3.2** (Nonparanormal distribution). A random vector $\boldsymbol{X} = (X_1, X_2, \ldots, X_d)^\top$ is nonparanormal, denoted by $NPN_d(\boldsymbol{\Sigma}^*; f_1, f_2, \ldots, f_d)$, if there exist monotone functions $f_1, f_2, \ldots, f_d$ such that $(f_1(X_1), f_2(X_2), \ldots, f_d(X_d))^\top \sim N_d(\mathbf{0}, \boldsymbol{\Sigma}^*)$.

Motivated by the extension from normal distribution to nonparanormal distribution, Liu et al. (2012b) furthermore proposed a semiparametric extension of elliptical distribution, which is called transelliptical distribution and defined as follows:

**Definition 3.3** (Transelliptical distribution). A random vector $\boldsymbol{X} = (X_1, X_2, \ldots, X_d)^\top \in \mathbb{R}^d$ is transelliptical, denoted by $TE_d(\boldsymbol{\Sigma}^*, \xi; f_1, f_2, \ldots, f_d)$, if there exists a set of monotone univariate functions $f_1, f_2, \ldots, f_d$ and a nonnegative random variable $\xi$, such that $(f_1(X_1), f_2(X_2), \ldots, f_d(X_d))^\top$ follows an elliptical distribution $EC_d(\boldsymbol{0}, \boldsymbol{\Sigma}^*, \xi)$.

It is worth noting that nonparanormal distribution is a special case of transelliptical distribution. In transelliptical graphical models, the latent precision matrix $\boldsymbol{\Theta}^* = (\boldsymbol{\Sigma}^*)^{-1}$ characterizes the conditional independence of the marginal random variables, i.e., $X_j$ and $X_k$ are independent conditioned on $X_\ell, \ell = 1, \ldots, d, \ell \neq j, k$ if and only if $\Theta^*_{jk} = 0$. This results in an undirected graph $G = (V, E)$, where $V$ contains nodes corresponding to the $d$ random variables in $\boldsymbol{X}$, and the edge set $E$ characterizes the conditional independence relationships among $X_1, \ldots, X_d$.

In the high dimensional regime, one often assumes $\boldsymbol{\Theta}^*$ is sparse, i.e., there are only a subset of variables that are correlated, which gives rise to a sparse graph. In order to estimate the sparse latent precision matrix $\boldsymbol{\Theta}^*$ and detect the correlation, we can first estimate the latent correlation matrix $\boldsymbol{\Sigma}^*$ based on Kendall's tau correlation coefficient (Kruskal, 1958), and then estimate $\boldsymbol{\Theta}^*$ using a rank-based estimator which we will introduce later.

## 3.1 Kendall's tau Statistic

In semiparametric setting, the Pearson's sample covariance matrix can be inconsistent in estimating $\boldsymbol{\Sigma}^*$. Given $n$ independent observations $\boldsymbol{X}_1, \ldots, \boldsymbol{X}_n$, where $\boldsymbol{X}_i = (X_{i1}, \ldots, X_{id})^\top \sim TE_d(\boldsymbol{\Sigma}^*, \xi; f_1, f_2, \ldots, f_d)$, Liu et al. (2012b) proposed a rank-based estimator, the Kendall's tau, to estimate $\boldsymbol{\Sigma}^*$, due to their invariance under monotonic marginal transformations. The Kendall's tau statistic is defined as

$$\widehat{\tau}_{jk} = \frac{2}{n(n-1)} \sum_{1 \leq i < i' \leq n} \text{sign}\left[(X_{ij} - X_{i'j})(X_{ik} - X_{i'k})\right]. \tag{3.1}$$

Liu et al. (2012b) showed that $\widehat{\tau}_{jk}$ is an unbiased estimator of $\tau_{jk}$, the population Kendall's tau statistic between $X_{ij}$ and $X_{ik}$. According to Han and Liu (2012), we have $\Sigma^*_{jk} = \sin(\pi/2 \cdot \tau_{jk})$, where $\Sigma^*_{jk}$ is Pearson's correlation between $X_{ij}$ and $X_{ik}$.

## 3.2 Rank-based Estimator

To define the rank-based estimator for latent precision matrix $\boldsymbol{\Theta}^* = (\boldsymbol{\Sigma}^*)^{-1}$, we first estimate the latent correlation matrix $\boldsymbol{\Sigma}^*$ by $\widehat{\boldsymbol{\Sigma}} = [\widehat{\Sigma}_{jk}] \in \mathbb{R}^{d \times d}$, where $\widehat{\Sigma}_{jk}$ is defined as:

$$\widehat{\Sigma}_{jk} = \begin{cases} \sin\left(\frac{\pi}{2}\widehat{\tau}_{jk}\right), & j \neq k, \\ 1, & j = k. \end{cases} \tag{3.2}$$

In the high dimensional regime, the latent precision matrix $\boldsymbol{\Theta}^*$ is often assumed to be sparse. We plug the estimated correlation matrix $\widehat{\boldsymbol{\Sigma}}$ into the CLIME estimator proposed in Cai et al. (2011) to approximate $\boldsymbol{\Theta}^*$. The main idea is to estimate $\boldsymbol{\Theta}^*$ in a column-by-column fashion. CLIME solves the following constrained problem

$$\widehat{\boldsymbol{\Theta}}' = \underset{\boldsymbol{\Theta}}{\operatorname{argmin}} \|\boldsymbol{\Theta}\|_{1,1}, \text{ subject to } \|\widehat{\boldsymbol{\Sigma}}\boldsymbol{\Theta} - \mathbf{I}\|_{\infty,\infty} \leq \lambda, \qquad (3.3)$$

where $\lambda > 0$ is a regularization parameter that sparsifies the estimator. To ensure that the estimator of $\boldsymbol{\Theta}^*$ is symmetric, we use the following symmetrization step to obtain the CLIME estimator $\widehat{\boldsymbol{\Theta}}$:

$$\widehat{\Theta}_{jk} = \widehat{\Theta}'_{jk} \mathbb{1}(|\widehat{\Theta}'_{jk}| \leq |\widehat{\Theta}'_{kj}|) + \widehat{\Theta}'_{kj} \mathbb{1}(|\widehat{\Theta}'_{jk}| > |\widehat{\Theta}'_{kj}|).$$

We refer to Cai et al. (2011) for more details. It is worth noting that (3.3) can be formulated as a linear programming, and solved column by column, and that CLIME is able to scale up to large datasets (Pang et al., 2014).

## 4 The Proposed Distributed Algorithms

In this section, we propose a distributed method for estimation and inference of the latent precision matrix in transelliptical graphical models. We begin with the problem setting of distributed estimation.

### 4.1 Problem Setting

Given $N$ i.i.d. observations from $d$-dimensional transelliptical distribution, i.e., $\boldsymbol{X}_i = (X_{i1}, ..., X_{id})^\top \sim TE_d(\boldsymbol{\Sigma}^*, \xi; f_1, \ldots, f_d), i = 1, \cdots, N$, we partition them into $m$ groups with each group of samples processed on one machine. The data matrix on each machine is denoted as $\mathbf{X}^{(l)} \in \mathbb{R}^{n_l \times d}, l = 1, \cdots, m$, where $m$ is the total number of machine and $n_l$ is the sample size on the $l$-th machine. Without loss of generality, we assume $n_1 = \ldots = n_m = n$ and $N = n \times m$. A naive proposal is to apply the CLIME method to the data on each machine to obtain $\widehat{\boldsymbol{\Theta}}^{(l)}$ and average them together. However, this is problematic in the high dimensional regime, since $\widehat{\boldsymbol{\Theta}}^{(l)}$'s are biased estimators, and directly averaging them will lead to an estimator that is far from the true latent precision matrix $\boldsymbol{\Theta}^*$. To overcome this problem, we plan to develop a distributed algorithm for graph estimation as is stated in the following section.

### 4.2 Distributed Latent Precision Matrix Estimation

The proposed algorithm is described in detail as follows. Firstly, on each machine we estimate the latent precision matrix $\widehat{\boldsymbol{\Theta}}^{(l)}$ based on $\mathbf{X}^{(l)}$ using the rank-based CLIME estimator in Section 3.2. Note that the CLIME estimator is a biased estimator of the latent precision matrix. Hence it is necessary to perform a debiasing step on $\widehat{\boldsymbol{\Theta}}^{(l)}$'s. In particular, we adopt the debiased estimator proposed in Jankova and van de Geer (2013), which takes the following form:

$$\widetilde{\boldsymbol{\Theta}}^{(l)} = 2\widehat{\boldsymbol{\Theta}}^{(l)} - \widehat{\boldsymbol{\Theta}}^{(l)}\widehat{\boldsymbol{\Sigma}}^{(l)}\widehat{\boldsymbol{\Theta}}^{(l)}, \qquad (4.1)$$

where $\widetilde{\boldsymbol{\Theta}}^{(l)}$ is the debiased estimator generated by the $l$-th machine. Note that $\widetilde{\boldsymbol{\Theta}}^{(l)}$'s are no longer sparse. After $\widetilde{\boldsymbol{\Theta}}^{(l)}$'s are obtained they are sent to the master machine and averaged, and we get

**Algorithm 1** Distributed Estimation of Transelliptical Graphical Models

**Require:** $\mathbf{X}^{(l)} \in \mathbb{R}^{n \times d}, l = 1, 2, \ldots, m$, hard thresholding parameter $t$.
**Ensure:** $\bar{\Theta}$
  **Workers:**
  Each worker computes $\widehat{\Sigma}_{jk}^{(l)}$ for all $j, k = 1, \ldots, d$, by (3.2) and (3.1);
  Each worker computes $\widehat{\Theta}^{(l)}$ via $\widehat{\Sigma}^{(l)}$ by (3.3);
  Each worker computes the unbiased estimator $\widetilde{\Theta}^{(l)}$ by (4.1);
  Each worker sends $\widetilde{\Theta}^{(l)}$ to the master machine
  **Master:**
  **while** waiting for $\widetilde{\Theta}^{(l)}$ sent from all workers **do**
    **if** received $\widetilde{\Theta}^{(l)}$ from all workers **then**
      Compute the final estimator $\check{\Theta}$ by (4.2).
    **end if**
  **end while**

$\bar{\Theta} := 1/m \sum_{l=1}^{m} \widetilde{\Theta}^{(l)}$. Since $\bar{\Theta}$ is not sparse either, we use a hard thresholding function $\mathrm{HT}(\cdot, t)$ to sparsify the averaged estimator, with $t$ being the threshold parameter:

$$\check{\Theta} := \mathrm{HT}(\bar{\Theta}, t), \quad \text{where} \quad \check{\Theta}_{jk} = \bar{\Theta}_{jk} \cdot \mathbb{1}(|\bar{\Theta}_{jk}| > t), \quad \text{for} \quad j, k = 1, \cdots, d, \quad (4.2)$$

where $\mathbb{1}(\cdot)$ is the indicator function, $\check{\Theta}$ is the sparsified averaged estimator, and $\mathrm{HT}(\cdot, t)$ means selecting the entries of which the absolute values are larger than $t$. The pseudo code of the algorithm is summarized in Algorithm 1.

We prove in our main theory that the distributed estimator $\check{\Theta}$ enjoys the same statistical properties as the estimator which is calculated based on combining the data from all the machines (Rothman et al., 2008; Ravikumar et al., 2011; Cai et al., 2011). We will address an important question that how to choose $m$, i.e., the number of machines, as $n$ grows large, providing a theoretical upper bound on $m$ such that the information loss of the proposed algorithm due to limited communication is negligible. Note that the above distributed estimation framework can be applied to precision matrix estimation in Gaussian graphical models (Meinshausen and Bühlmann, 2006; Yuan and Lin, 2007; Banerjee et al., 2008; Friedman et al., 2008; Jankova and van de Geer, 2013) straightforwardly, because the Gaussian graphical model is a special case of the Gaussian copula graphical model, where all the marginal monotone transformations are chosen to be identity functions.

Table 1: Time complexity comparison between centralized and distributed estimators.

| Algorithm | Centralized Estimator | Distributed Estimator |
| --- | --- | --- |
| Time Complexity | $O(d^2 N^2 + d^3)$ | $O(d^2 N^2 / m^2 + d^3)$ |

Since we only need to send $\widetilde{\Theta}^{(l)}, l = 1, \ldots, m$ once to the master machine, our algorithm is very communication-efficient. The time complexity comparison between the distributed estimator and the centralized estimator is summarized in Table 1. Specifically, for the centralized method, we need to calculate Kendall's tau statistics for each entry of the latent correlation matrix, and each calculation is performed on $N$ samples, whose time complexity is $O(N^2)$. So the total time complexity to compute the latent correlation matrix is $O(d^2 N^2)$. The time complexity of the estimation for the latent precision matrix is $O(d^3)$. Hence the total time complexity of centralized estimation is $O(d^2 N^2 + d^3)$. For the distributed method, the total time complexity for calculating Kendall's tau statistics on each machine is reduced to $O(d^2 N^2 / m^2)$. The time complexity of CLIME is still $O(d^3)$ and the debiasing step also needs $O(d^3)$ time complexity due to $d \times d$ matrix multiplication. This leads to the total time complexity of $O(d^2 N^2 / m^2 + d^3)$. Apparently, the proposed distributed method saves a lot of time compared with the centralized method.

### 4.3 Distributed Asymptotic Inference

Apart from the estimation of the latent precision matrix, we prove the asymptotic property of the averaged debiased estimator of the latent precision matrix for transelliptical graphical models. Based on the asymptotic property, a test statistic and an inference procedure are also proposed. Recall the problem setting in Section 4.1, $\mathbf{X}^{(l)} \in \mathbb{R}^{n \times d}$ is the data matrix on the $l$-th machine. By the definition of Kendall's tau estimator in Section 3, we have

$$\widehat{\tau}_{pq}^{(l)} = \frac{2}{n(n-1)} \sum_{1 \leq i < i' \leq n} \mathrm{sign}\left[\left(X_{ip}^{(l)} - X_{i'p}^{(l)}\right)\left(X_{iq}^{(l)} - X_{i'q}^{(l)}\right)\right], \tag{4.3}$$

for any $p, q = 1, \ldots, d$. In Section 3, we know that $\widehat{\tau}_{pq}^{(l)}$ is an unbiased estimator of $\tau_{pq} = 2/\pi \arcsin(\Sigma_{pq}^*)$. The latent correlation matrix $\mathbf{\Sigma}^* = [\Sigma_{pq}^*]$ can be estimated by $\widehat{\Sigma}_{pq}^{(l)} = \sin\left(\pi \widehat{\tau}_{pq}^{(l)} / 2\right)$ on the $l$-th machine. By definition, $\widehat{\tau}_{pq}^{(l)}$ is a $U$-statistic, and thus we employ Hájek's projection method (Hoeffding, 1948) for $U$ statistics in our analysis. We first define the following notations on the $l$-th group of samples. To ease the notation, we just leave out the index $l$ when no confusion arises.

$$\begin{aligned}
h_{pq}^{ii'} &= \mathrm{sign}\big((X_{ip} - X_{i'p})(X_{iq} - X_{i'q})\big) - \tau_{pq}, \\
h_{pq}^{ii'|i} &= \mathbb{E}[h_{pq}^{ii'} | \mathbf{X}_{i*}] \\
&= \left[\mathbb{P}\Big((X_{ip} - X_{i'p})(X_{iq} - X_{i'q}) > 0 | \mathbf{X}_{i*}\Big) - \mathbb{P}\Big((X_{ip} - X_{i'p})(X_{iq} - X_{i'q}) < 0 | \mathbf{X}_{i*}\Big)\right] - \tau_{pq}, \\
h_{pq}^{i} &= \frac{1}{n-1} \sum_{i' \neq i} h_{pq}^{ii'|i}, \\
w_{pq}^{ii'} &= h_{pq}^{ii'} - h_{pq}^{ii'|i} - h_{pq}^{ii'|i'},
\end{aligned} \tag{4.4}$$

where $\mathbf{X}_{i*}$ is the $i$-th row of the data matrix $\mathbf{X}^{(l)}$ on the $l$-th machine. Note that $h_{pq}^{ii'|i}$ is still a random variable depending on $\mathbf{X}_{i*}$. We then define a matrix $\mathbf{M}^i$ for $i = 1, \ldots, n$ where

$$M_{pq}^i = \pi \cos\left(\tau_{pq} \frac{\pi}{2}\right) h_{pq}^i. \tag{4.5}$$

By the definition of $h_{pq}^i$, we have $\mathbb{E}[M_{pq}^i] = 0$. Specifically, for any $j, k = 1, \ldots, d$, we prove in our main theory part that

$$\frac{\sqrt{N}}{m} \sum_{l=1}^{m} \frac{\bar{\Theta}_{jk} - \Theta_{jk}^*}{\sigma_{jk,l}} \rightsquigarrow -\frac{\sqrt{N}}{m} \sum_{l=1}^{m} \frac{\Theta_{*j}^{*\top} \widehat{\Sigma}^{(l)} \Theta_{*k}^* - \Theta_{jk}^*}{\sigma_{jk,l}} \rightsquigarrow N(0,1), \tag{4.6}$$

where $\sigma_{jk,l}^2 := 1/n \sum_{i=1}^{n} \text{Var}(\Theta_{*j}^{*\top} \mathbf{M}^i \Theta_{*k}^*)$ is the variance calculated on the $l$-th machine, and $\mathbf{M}^i$ is defined in (4.5) based on data matrix $\mathbf{X}^{(l)}$. To obtain the confidence interval, we need to find a consistent estimator of $\sigma_{jk,l}^2$. To this end we define the following matrices $\widehat{\mathbf{M}}^i$:

$$\widehat{h}_{pq}^i = \frac{1}{n-1} \sum_{i' \neq i} \text{sign}\big((X_{ip} - X_{i'p})(X_{iq} - X_{i'q})\big) - \widehat{\tau}_{pq},$$

$$\widehat{M}_{pq}^i = \pi \cos\left(\frac{\pi}{2} \widehat{\tau}_{pq}\right) \widehat{h}_{pq}^i. \tag{4.7}$$

We then define the estimator of $\sigma_{jk,l}^2$ as $\widehat{\sigma}_{jk,l}^2 := 1/n \sum_{i=1}^{n} \big(\widehat{\Theta}_{*j}^{(l)\top} \widehat{\mathbf{M}}^i \widehat{\Theta}_{*k}^{(l)}\big)^2$, where $\widehat{\mathbf{M}}^i$ is calculated based on samples $\mathbf{X}_{(l-1)n+1}, \ldots, \mathbf{X}_{ln}$ for $(l-1)n + 1 \leq i \leq ln$. We prove in our main theory part that $\widehat{\sigma}_{jk,l}^2 \xrightarrow{p} \sigma_{jk,l}^2$, which implies that $\widehat{\sigma}_{jk,l}^2$ is a consistent estimator of the variance. Plugging $\widehat{\sigma}_{jk,l}^2$ into (4.6), by Slutsky's theorem we obtain

$$\frac{\sqrt{N}}{m} \sum_{l=1}^{m} \frac{\bar{\Theta}_{jk} - \Theta_{jk}^*}{\widehat{\sigma}_{jk,l}} \rightsquigarrow N(0,1).$$

Thus we obtain $1 - \alpha$ confidence interval of $\Theta_{jk}^*$

$$\left[\bar{\Theta}_{jk} - \frac{\sqrt{m} u_{1-\alpha/2}}{\sqrt{n}} \left(\sum_{l=1}^{m} \frac{1}{\widehat{\sigma}_{jk,l}}\right)^{-1}, \bar{\Theta}_{jk} + \frac{\sqrt{m} u_{1-\alpha/2}}{\sqrt{n}} \left(\sum_{l=1}^{m} \frac{1}{\widehat{\sigma}_{jk,l}}\right)^{-1}\right],$$

where $\widehat{\sigma}_{jk,l}^2 = 1/n \sum_{i=1}^{n} \big(\widehat{\Theta}_{*j}^{(l)\top} \widehat{\mathbf{M}}^i \widehat{\Theta}_{*k}^{(l)}\big)^2$ and $u_{1-\alpha/2}$ is the $1 - \alpha/2$ quantile of standard normal distribution. Under the null hypothesis $H_0 : \Theta_{jk}^* = 0$, our test statistic is given by

$$\widehat{U}_n = \frac{\sqrt{N}}{m} \sum_{l=1}^{m} \frac{\bar{\Theta}_{jk}}{\widehat{\sigma}_{jk,l}} = \frac{N}{m^{3/2}} \sum_{l=1}^{m} \frac{\bar{\Theta}_{jk}}{\sqrt{\sum_{i=1}^{n} \big(\widehat{\Theta}_{*j}^{(l)\top} \widehat{\mathbf{M}}^i \widehat{\Theta}_{*k}^{(l)}\big)^2}}. \tag{4.8}$$

Then the Wald test with significant level $\alpha$ is given by

$$\Psi(\alpha) = \mathbb{1}\big(|\widehat{U}_n| > u_{1-\alpha/2}\big). \tag{4.9}$$

The null hypothesis is rejected when $\Psi(\alpha) = 1$, and the associated $p$-value is given by $p = 2\big(1 - \Phi(|\widehat{U}_n|)\big)$, where $\Phi(\cdot)$ is the cumulative distribution function of standard normal distribution. More details are discussed in Section 5.2.

# 5  Main Theory

In this section, we present our main theoretical results. We start with some assumptions, which are required throughout our analysis.

**Assumption 5.1.** For simplicity, we consider the exactly sparse matrix class. For constants $M > 0$ and $s > 0$, suppose the true latent precision matrix $\boldsymbol{\Theta}^*$ belongs to the following class of matrices

$$\mathcal{U}(s, M) = \left\{ \boldsymbol{\Theta} \in \mathbb{R}^{d \times d} : \boldsymbol{\Theta} \succ \boldsymbol{0}, \|\boldsymbol{\Theta}\|_1 \leq M, \max_{1 \leq j \leq d} \sum_{k=1}^{d} \mathbb{1}(\Theta_{jk} \neq 0) \leq s \right\}, \tag{5.1}$$

where $\boldsymbol{\Theta} = [\Theta_{jk}]$, $\boldsymbol{\Theta} \succ \boldsymbol{0}$ indicates that $\boldsymbol{\Theta}$ is symmetric and positive definite.

The above assumption has been widely made in the literature of Gaussian graphical model estimation (Cai et al., 2011) as well as semiparametric graphical model estimation (Liu et al., 2012b).

Without loss of generality, we assume that the eigenvalues of $\boldsymbol{\Sigma}^* = (\boldsymbol{\Theta}^*)^{-1}$ are bounded. More specifically, we have the following assumption.

**Assumption 5.2.** There exists a constant $\nu > 0$ such that $1/\nu \leq \lambda_{\min}(\boldsymbol{\Sigma}^*) \leq \lambda_{\max}(\boldsymbol{\Sigma}^*) \leq \nu$. In addition, there exists a constant $K_{\boldsymbol{\Sigma}^*} > 0$, such that $\|\boldsymbol{\Sigma}^*\|_\infty \leq K_{\boldsymbol{\Sigma}^*}$.

The first part of Assumption 5.2 requires that the smallest eigenvalue of the latent correlation matrix $\boldsymbol{\Sigma}^*$ is bounded below from zero, and its largest eigenvalue is finite. Assumption 5.2 immediately implies that $1/\nu \leq \lambda_{\min}(\boldsymbol{\Theta}^*) \leq \lambda_{\max}(\boldsymbol{\Theta}^*) \leq \nu$. This assumption is commonly imposed in the literature for the analysis of Gaussian graphical models (Ravikumar et al., 2011) and nonparanormal graphical models (Liu et al., 2009).

## 5.1  Main Results for Distributed Latent Precision Matrix Estimation

Now we are ready to present our main results. Recall that we obtain $\widehat{\boldsymbol{\Theta}}^{(l)}$ on each machine by CLIME method, and then debias them as follows: $\widetilde{\boldsymbol{\Theta}}^{(l)} = 2\widehat{\boldsymbol{\Theta}}^{(l)} - \widehat{\boldsymbol{\Theta}}^{(l)} \widehat{\boldsymbol{\Sigma}}^{(l)} \widehat{\boldsymbol{\Theta}}^{(l)}$. After we get the debiased estimators, we average them on a central machine as $\bar{\boldsymbol{\Theta}} = 1/m \sum_{l=1}^{m} \widetilde{\boldsymbol{\Theta}}^{(l)}$. In the end, we use the thresholding function in (4.2) to get the sparsified averaged estimator $\check{\boldsymbol{\Theta}} = \text{HT}(\bar{\boldsymbol{\Theta}}, t)$.

We show in Theorem 5.3 that the sparsified averaged estimator $\check{\boldsymbol{\Theta}}$ enjoys the same statistical properties as the estimator that is based on combining the data from all the machines and followed by CLIME estimation process, i.e., the centralized estimator.

**Theorem 5.3.** Under Assumptions 5.1 and 5.2, if the thresholding parameter is chosen to be $t = 4M^2\sqrt{\log d/N} + M^2\sqrt{d/(N(n-1))} + CM^4 K_{\boldsymbol{\Sigma}^*} sm \log d/N$, then with probability at least $1 - 76/d$, the estimator $\check{\boldsymbol{\Theta}}$ satisfies

$$\|\check{\boldsymbol{\Theta}} - \boldsymbol{\Theta}^*\|_{\infty,\infty} \leq 8M^2\sqrt{\frac{\log d}{N}} + 2M^2\sqrt{\frac{d}{N(n-1)}} + 2CM^4 K_{\boldsymbol{\Sigma}^*} \frac{sm \log d}{N}, \tag{5.2}$$

$$\|\check{\boldsymbol{\Theta}} - \boldsymbol{\Theta}^*\|_2 \leq 8M^2 s\sqrt{\frac{\log d}{N}} + 2M^2 s\sqrt{\frac{d}{N(n-1)}} + 2CM^4 K_{\boldsymbol{\Sigma}^*} \frac{s^2 m \log d}{N}, \tag{5.3}$$

$$\|\check{\boldsymbol{\Theta}} - \boldsymbol{\Theta}^*\|_F \leq 8M^2 \sqrt{\frac{ds \log d}{N}} + 2M^2 \sqrt{\frac{sd^2}{N(n-1)}} + 2CM^4 K_{\boldsymbol{\Sigma}^*} \frac{ms^{3/2}\sqrt{d} \log d}{N}, \tag{5.4}$$

where $C > 0$ is an absolute constant.

From Theorem 5.3, we can see that the estimation error bounds of the proposed distributed estimator consist of two parts. Take the infinity norm $\|\check{\Theta} - \Theta^*\|_{\infty,\infty}$ for example, the first part is the first term $O(\sqrt{\log d/N})$ in (5.2), which attains the same rate as the centralized estimator. The second and third terms in (5.2) are the second part, introduced by the information loss in data distribution. Based on Theorem 5.3, we immediately have the following corollary specifying the scaling of $m$.

**Corollary 5.4.** Under the same assumptions as in Theorem 5.3, if the number of machines $m$ satisfies

$$m \lesssim \min\left\{\frac{N \log d}{d}, \sqrt{\frac{N}{s^2 \log d}}\right\}, \tag{5.5}$$

then with probability at least $1 - 76/d$, the estimator $\check{\Theta}$ in (4.2) satisfies

$$\|\check{\Theta} - \Theta^*\|_{\infty,\infty} \leq (9 + 2CM^2 K_{\Sigma^*})M^2 \sqrt{\frac{\log d}{N}}, \quad \|\check{\Theta} - \Theta^*\|_2 \leq (9 + 2CM^2 K_{\Sigma^*})M^2 s \sqrt{\frac{\log d}{N}},$$

$$\|\check{\Theta} - \Theta^*\|_F \leq (9 + 2CM^2 K_{\Sigma^*})M^2 \sqrt{\frac{sd \log d}{N}},$$

where $C > 0$ is an absolute constant.

**Remark 5.5.** It is well known that distributed estimation may cause information loss and often leads to a worse estimation error bound. Nevertheless, we prove in Corollary 5.4 that, if the number of machines $m$ is chosen properly as in (5.5), our distributed estimator is able to attain the same convergence rate as the centralized estimator. More specifically, our distributed method requires $m \lesssim \min\{N \log d/d, \sqrt{N/(s^2 \log d)}\}$ in order to attain the same rate as centralized method. For the first term, it is reasonable to require $m \lesssim N \log d/d$ in distributed setting, where the total sample size $N$ is often larger than $d$ (otherwise distributed estimation may not be necessary). And the second term is a very common scaling condition for number of machines in distributed learning, such as distributed sparse regression (Lee et al., 2015; Battey et al., 2015) and distributed linear discriminant analysis (Tian and Gu, 2016). Very recently, Arroyo and Hou (2016) also reaches the same condition that $m \leq n/(s^2 \log d)$ for the scaling of number of machines for the distributed Gaussian graphical model, which is a special case of our distributed transelliptical graphical model.

In addition, we are able to achieve the model selection consistency by assuming that the nonzero entries of $\Theta^*$ are large enough. More specifically, we have the following theorem.

**Theorem 5.6.** Under the same conditions as Theorem 5.3, let $\theta_{\min} = \min_{(j,k)\in\text{supp}(\Theta^*)} |\Theta^*_{jk}|$. If $\theta_{\min}$ satisfies

$$\theta_{\min} > 8M^2 \sqrt{\frac{\log d}{N}} + 2M^2 \sqrt{\frac{d}{N(n-1)}} + 2C \frac{sm \log d}{N},$$

where $C > 0$ is a constant, then $\text{supp}(\check{\Theta}) = \text{supp}(\Theta^*)$ holds with probability at least $1 - 76/d$.

Similar to Corollary 5.4, if the number of machines $m$ is chosen properly, Theorem 5.6 immediately implies the following corollary, which states that the distributed estimator can attain the model selection consistency under the same conditions as centralized estimator.

**Corollary 5.7.** Under the same conditions as in Theorem 5.3, we choose the number of machines $m$ as

$$m \lesssim \min\left\{\frac{N \log d}{d}, \sqrt{\frac{N}{s^2 \log d}}\right\}, \tag{5.6}$$

and $\boldsymbol{\Theta}^*$ satisfies

$$\theta_{\min} \geq 16M^2 \sqrt{\frac{\log d}{N}},$$

where $\theta_{\min} = \min_{(j,k) \in \text{supp}(\boldsymbol{\Theta}^*)} |\Theta^*_{jk}|$. Then with probability at least $1 - 76/d$ we have $\text{supp}(\check{\boldsymbol{\Theta}}) = \text{supp}(\boldsymbol{\Theta}^*)$.

**Remark 5.8.** Corollary 5.7 indicates that our distributed estimator can achieve model selection consistency under the condition that $\theta_{\min} \gtrsim M^2 \sqrt{\log d/N}$ when the number of machines $m$ is chosen properly. This matches the results of model selection consistency for centralized estimator in Cai et al. (2011), which also requires $\theta_{\min} \gtrsim M^2 \sqrt{\log d/N}$. Ravikumar et al. (2011) also proved the model selection consistency for graphical Lasso centralized estimator with the condition that $\theta_{\min} \gtrsim \sqrt{\log d/N}$. However, the graphical Lasso estimator additionally requires the irrepresentable condition, which is a particularly stringent assumption.

## 5.2 Main Results for Distributed Asymptotic Inference

So far, we have proved that our distributed estimator can attain the same rate of convergence as the centralized estimator. Now we consider the asymptotic properties of the distributed estimator $\bar{\boldsymbol{\Theta}}$ for transelliptical graphical models. Here we use the averaged debiased estimator to conduct the hypothesis test since the hard thresholding process is not essential to find the confidence interval for the true precision matrix. Recall the notations in Section 4.3, we have the following theorem:

**Theorem 5.9.** Suppose we have $\boldsymbol{X}_1, \ldots, \boldsymbol{X}_N$ i.i.d.$\sim TE_d(\boldsymbol{\Sigma}^*, \xi; f_1, \ldots, f_d)$. Under Assumptions 5.1 and 5.2, we further assume that $s \log d/\sqrt{n} = o(1)$. Recall the matrix $\mathbf{M}^i$ defined in (4.5) which is calculated based on samples $\boldsymbol{X}_{(l-1)n+1}, \ldots, \boldsymbol{X}_{ln}$ for $(l-1)n+1 \leq i \leq ln$, and suppose we have

$$\text{Var}(\boldsymbol{\Theta}^{*\top}_{*j} \mathbf{M}^i \boldsymbol{\Theta}^*_{*k}) \geq \rho_{\min} \|\boldsymbol{\Theta}^*_{*j}\|_2^2 \cdot \|\boldsymbol{\Theta}^*_{*k}\|_2^2$$

holds for any $j, k = 1, \ldots, d$, where $\boldsymbol{\Theta}^* = (\boldsymbol{\Sigma}^*)^{-1}$ and $\rho_{\min} > 0$ is a constant. Then we have

$$\frac{\sqrt{N}}{m} \sum_{l=1}^{m} \frac{\bar{\Theta}_{jk} - \Theta^*_{jk}}{\sigma_{jk,l}} \rightsquigarrow -\frac{\sqrt{N}}{m} \sum_{l=1}^{m} \frac{\boldsymbol{\Theta}^{*\top}_{*j} \widehat{\boldsymbol{\Sigma}}^{(l)} \boldsymbol{\Theta}^*_{*k} - \Theta^*_{jk}}{\sigma_{jk,l}} \rightsquigarrow N(0,1),$$

where $\sigma_{jk,l}^2 := 1/n \sum_{i=(l-1)n+1}^{ln} \text{Var}(\boldsymbol{\Theta}^{*\top}_{*j} \mathbf{M}^i \boldsymbol{\Theta}^*_{*k})$.

**Remark 5.10.** By the conditions of Theorem 5.9 it can be seen that the scaling for the asymptotic normality to hold is $n \gtrsim (s \log d)^2$. This scaling condition is identical to that of the centralized asymptotic inference methods for Gaussian (Jankova and van de Geer, 2013; Liu et al., 2013), nonparanormal (Gu et al., 2015) and transelliptical graphical models (Barber and Kolar, 2015), and has been proved to be optimal in Ren et al. (2015).

Since $\boldsymbol{\Theta}^*$ is unknown in practice and so is the variance term $\sigma_{jk,l}^2$, in order to make $\sqrt{N}/m \sum_{l=1}^{m} (\bar{\Theta}_{jk} - \Theta_{jk}^*)/\sigma_{jk,l}$ a valid statistic, we need to find a consistent estimator of $\sigma_{jk,l}^2$. The following proposition gives a consistent estimator for $\sigma_{jk,l}^2$.

**Proposition 5.11.** Under Assumptions 5.1 and 5.2, suppose that for any $j, k = 1, \ldots, d$,

$$M^4 \cdot s\sqrt{\log d/n} = o(1), \tag{5.7}$$

$$M^4 \cdot \sqrt{\log(nd)/n} = o(1). \tag{5.8}$$

Then we have that $\widehat{\sigma}_{jk,l}^2 \xrightarrow{p} \sigma_{jk,l}^2$, where $\widehat{\sigma}_{jk,l}^2 = 1/n \sum_{i=(l-1)n+1}^{ln} (\widehat{\boldsymbol{\Theta}}_{*j}^{(l)\top} \widehat{\mathbf{M}}^i \widehat{\boldsymbol{\Theta}}_{*k}^{(l)})^2$, and $\widehat{\mathbf{M}}^i$ is defined in (4.7).

Due to the consistency of $\widehat{\sigma}_{jk,l}^2$, we replace $\sigma_{jk,l}^2$ in Theorem 5.9 with the plugin estimator $\widehat{\sigma}_{jk,l}^2 = 1/n \sum_{i=(l-1)n+1}^{ln} (\widehat{\boldsymbol{\Theta}}_{*j}^{(l)\top} \widehat{\mathbf{M}}^i \widehat{\boldsymbol{\Theta}}_{*k}^{(l)})^2$. Then by Slutsky's theorem we have $\sqrt{N}/m \sum_{l=1}^{m} (\bar{\Theta}_{jk} - \Theta_{jk}^*)/\widehat{\sigma}_{jk,l} \rightsquigarrow N(0,1)$. Under $H_0 : \Theta_{jk}^* = 0$, the test statistic $\widehat{U}_n$ is given by (4.8). Then we have the following corollary:

**Corollary 5.12.** Under the same assumptions in Theorem 5.9, under $H_0 : \Theta_{jk}^* = 0$, we have

$$\lim_{n \to \infty} \sup_{t \in \mathbb{R}} |\mathbb{P}(\widehat{U}_n \leq t) - \Phi(t)| = 0, \tag{5.9}$$

where $\Phi(\cdot)$ is the cumulative distribution function of standard normal distribution.

**Remark 5.13.** Under the significance level $\alpha = 0.05$, the null hypothesis is rejected if $|\widehat{U}_n| > u_{1-\alpha/2}$, and the associated $p$-value is given by $p = 2(1 - \Phi(|\widehat{U}_n|))$.

# 6  Proofs of The Main Theory

In this section, we present the proofs of the main theorems in Section 5. The statements and proofs of several technical lemmas are in Appendix.

## 6.1  Proof of Theorem 5.3

In order to prove Theorem 5.3, we need the following technical lemma.

**Lemma 6.1.** Under Assumptions 5.1 and 5.2, the debiased estimator on each machine is given by $\widetilde{\boldsymbol{\Theta}}^{(l)} = 2\widehat{\boldsymbol{\Theta}}^{(l)} - \widehat{\boldsymbol{\Theta}}^{(l)} \widehat{\boldsymbol{\Sigma}}^{(l)} \widehat{\boldsymbol{\Theta}}^{(l)}$ for $l = 1, \ldots, d$. Then with probability at least $1 - 76/d$, the average of debiased estimators $\bar{\boldsymbol{\Theta}} = 1/m \sum_{l=1}^{m} \widetilde{\boldsymbol{\Theta}}^{(l)}$ satisfies

$$\|\bar{\boldsymbol{\Theta}} - \boldsymbol{\Theta}^*\|_{\infty,\infty} \leq 4M^2 \sqrt{\frac{\log d}{N}} + M^2 \sqrt{\frac{d}{N(n-1)}} + CM^4 K_{\Sigma^*} \frac{sm \log d}{N},$$

where $C > 0$ is an absolute constant.

*Proof of Theorem 5.3.* By the choice of the thresholding parameter, we have $\|\bar{\boldsymbol{\Theta}} - \boldsymbol{\Theta}^*\|_{\infty,\infty} \leq t$. Recall the definition of hard thresholding function, and we have that $\text{supp}(\check{\boldsymbol{\Theta}})$ is the set of index $(j, k)$ such that $\bar{\Theta}_{jk} > t$. Thus, for any $(j,k) \in \text{supp}(\check{\boldsymbol{\Theta}})$, we have $\check{\Theta}_{jk} = \bar{\Theta}_{jk}$; for $(j,k) \notin \text{supp}(\check{\boldsymbol{\Theta}})$, we have $|\check{\Theta}_{jk} - \bar{\Theta}_{jk}| = |0 - \bar{\Theta}_{jk}| \leq t$. Therefore, we obtain $\|\check{\boldsymbol{\Theta}} - \bar{\boldsymbol{\Theta}}\|_{\infty,\infty} \leq t$. By triangle inequality,

$$\|\check{\boldsymbol{\Theta}} - \boldsymbol{\Theta}^*\|_{\infty,\infty} \leq \|\check{\boldsymbol{\Theta}} - \bar{\boldsymbol{\Theta}}\|_{\infty,\infty} + \|\bar{\boldsymbol{\Theta}} - \boldsymbol{\Theta}^*\|_{\infty,\infty}$$
$$\leq t + t = 8M^2\sqrt{\frac{\log d}{N}} + 2M^2\sqrt{\frac{d}{N(n-1)}} + 2CM^4 K_{\boldsymbol{\Sigma}^*}\frac{sm\log d}{N}$$

holds with probability at least $1 - 76/d$, where the second inequality follows from Lemma 6.1. By Assumption 5.1, which states that $\boldsymbol{\Theta}^* \in \mathcal{U}(s, M)$, we have the row sparsity claim: $\|\boldsymbol{\Theta}^*\|_{\infty,0} = \max_j \sum_{k=1}^d \mathbb{1}(\Theta_{jk} \neq 0) \leq s$. When $t > \|\bar{\boldsymbol{\Theta}} - \boldsymbol{\Theta}^*\|_{\infty,\infty}$, we have $|\bar{\Theta}_{jk}| < t$ and $\check{\Theta}_{jk} = 0$ whenever $\Theta^*_{jk} = 0$ due to the thresholding function, which implies that $\text{supp}(\check{\boldsymbol{\Theta}}) \subset \text{supp}(\boldsymbol{\Theta}^*)$. Therefore we obtain $\|\check{\boldsymbol{\Theta}} - \boldsymbol{\Theta}^*\|_{\infty,0} \leq s$. By properties of matrix norms, we have $\|\boldsymbol{\Theta}\|_2^2 \leq \|\boldsymbol{\Theta}\|_1 \cdot \|\boldsymbol{\Theta}\|_\infty$ and $\|\boldsymbol{\Theta}\|_F \leq \sqrt{sd}\|\boldsymbol{\Theta}\|_{\infty,\infty}$. And for symmetric matrix $\boldsymbol{\Theta}$, we have $\|\boldsymbol{\Theta}\|_1 = \|\boldsymbol{\Theta}\|_\infty$. Then with probability at least $1 - 76/d$, we have

$$\|\check{\boldsymbol{\Theta}} - \boldsymbol{\Theta}^*\|_2 \leq \|\check{\boldsymbol{\Theta}} - \boldsymbol{\Theta}^*\|_1 \leq s \cdot \|\check{\boldsymbol{\Theta}} - \boldsymbol{\Theta}^*\|_{\infty,\infty}$$
$$\leq 8M^2 s\sqrt{\frac{\log d}{N}} + 2M^2 s\sqrt{\frac{d}{N(n-1)}} + 2CM^4 K_{\boldsymbol{\Sigma}^*}\frac{s^2 m\log d}{N},$$
$$\|\check{\boldsymbol{\Theta}} - \boldsymbol{\Theta}^*\|_F \leq \sqrt{ds}\|\check{\boldsymbol{\Theta}} - \boldsymbol{\Theta}^*\|_{\infty,\infty}$$
$$\leq 8M^2\sqrt{\frac{ds\log d}{N}} + 2M^2\sqrt{\frac{sd^2}{N(n-1)}} + 2CM^4 K_{\boldsymbol{\Sigma}^*}\frac{ms^{3/2}\sqrt{d}\log d}{N},$$

where $C$ is an absolute constant. $\square$

### 6.2 Proof of Theorem 5.6

Next we prove the model selection consistency.

*Proof of Theorem 5.6.* As stated in Theorem 5.3, we choose $t = 4M^2\sqrt{\log d/N} + M^2\sqrt{d/(N(n-1))} + C'sm\log d/N$. For any $(j,k) \notin \text{supp}(\boldsymbol{\Theta}^*)$, by the result in Lemma 6.1, we have that $|\bar{\Theta}_{jk} - \Theta^*_{jk}| = |\bar{\Theta}_{jk}| \leq t$. However, since $\check{\boldsymbol{\Theta}}$ is hard thresholded by choosing the entries of $\bar{\boldsymbol{\Theta}}$ that are larger than $t$, we get the conclusion that $\check{\Theta}_{jk} = 0$. It follows that $\text{supp}(\check{\boldsymbol{\Theta}}) \subseteq \text{supp}(\boldsymbol{\Theta}^*)$.

On the other hand, for any $(j,k) \notin \text{supp}(\check{\boldsymbol{\Theta}})$, similarly we have $|\check{\Theta}_{jk} - \Theta^*_{jk}| = |\Theta^*_{jk}| \leq 2t$. However, by assumption we have $\min_{(j,k)\in\text{supp}(\boldsymbol{\Theta}^*)} |\Theta^*_{jk}| > 2t$. Thus, we have $\Theta^*_{jk} = 0$ for any $(j,k) \notin \text{supp}(\check{\boldsymbol{\Theta}})$. It follows that $\text{supp}(\boldsymbol{\Theta}^*) \subseteq \text{supp}(\check{\boldsymbol{\Theta}})$. And finally, we obtain $\text{supp}(\check{\boldsymbol{\Theta}}) = \text{supp}(\boldsymbol{\Theta}^*)$. $\square$

### 6.3 Proof of Theorem 5.9

Now we prove the asymptotic results for distributed inference.

*Proof of Theorem 5.9.* We first define $\mathbf{W}^{(l)} = \widehat{\boldsymbol{\Sigma}}^{(l)} - \boldsymbol{\Sigma}^*$ and thus we have

$$\widetilde{\boldsymbol{\Theta}}^{(l)} - \boldsymbol{\Theta}^* = 2\widehat{\boldsymbol{\Theta}}^{(l)} - \widehat{\boldsymbol{\Theta}}^{(l)}\widehat{\boldsymbol{\Sigma}}^{(l)}\widehat{\boldsymbol{\Theta}}^{(l)} - \boldsymbol{\Theta}^* = -\boldsymbol{\Theta}^*\mathbf{W}^{(l)}\boldsymbol{\Theta}^* + \text{Rem}^{(l)}, \tag{6.1}$$

where $\text{Rem}^{(l)} = -(\widehat{\boldsymbol{\Theta}}^{(l)} - \boldsymbol{\Theta}^*)\mathbf{W}^{(l)}\boldsymbol{\Theta}^* - (\widehat{\boldsymbol{\Theta}}^{(l)}\widehat{\boldsymbol{\Sigma}}^{(l)} - \mathbf{I})(\widehat{\boldsymbol{\Theta}}^{(l)} - \boldsymbol{\Theta}^*)$ for $l = 1, \ldots, m$. For the average of debiased estimators $\bar{\boldsymbol{\Theta}}$, by the definition in (6.1) we have

$$\sqrt{n}(\bar{\Theta}_{jk} - \Theta^*_{jk}) = \frac{\sqrt{n}}{m}\sum_{l=1}^{m}\left(\widetilde{\Theta}^{(l)}_{jk} - \Theta^*_{jk}\right) = \underbrace{-\frac{\sqrt{n}}{m}\sum_{l=1}^{m}\left(\boldsymbol{\Theta}^*_{j*}\widehat{\boldsymbol{\Sigma}}^{(l)}\boldsymbol{\Theta}^*_{*k} - \Theta^*_{jk}\right)}_{I_{01}} + \underbrace{\frac{\sqrt{n}}{m}\sum_{l=1}^{m}\left[\text{Rem}^{(l)}\right]_{jk}}_{I_{02}}.$$

We denote $Z^{(l)}_{jk} = \sqrt{n}\big(\widetilde{\Theta}^{(l)}_{jk} - \Theta^*_{jk}\big)/\sigma_{jk}$ and $Z_{jk} = 1/m\sum_{l=1}^{m}Z^{(l)}_{jk} = \sqrt{n}(\bar{\Theta}_{jk} - \Theta^*_{jk})/\sigma_{jk}$. Note that we have $\widetilde{\Theta}^{(l)}_{jk} - \Theta^*_{jk} = \big(\boldsymbol{\Theta}^*_{j*}\widehat{\boldsymbol{\Sigma}}^{(l)}\boldsymbol{\Theta}^*_{*k} - \Theta^*_{jk}\big) + \left[\text{Rem}^{(l)}\right]_{jk}$. We will divide our proof into two part: to show that $I_{02} = o_p(1)$ and to show that $I_{01}$ is asymptotic normal.

**Term $I_{02}$:** We show that the $\ell_{\infty,\infty}$-norm of the term $\text{Rem}^{(l)}$ is asymptotically small. By triangle's inequality and Hölder's inequality we have

$$\|\text{Rem}^{(l)}\|_{\infty,\infty} \leq \|(\widehat{\boldsymbol{\Theta}}^{(l)} - \boldsymbol{\Theta}^*)\mathbf{W}^{(l)}\boldsymbol{\Theta}^*\|_{\infty,\infty} + \|(\widehat{\boldsymbol{\Theta}}^{(l)}\widehat{\boldsymbol{\Sigma}}^{(l)} - \mathbf{I})(\widehat{\boldsymbol{\Theta}}^{(l)} - \boldsymbol{\Theta}^*)\|_{\infty,\infty}$$
$$\leq \|\widehat{\boldsymbol{\Theta}}^{(l)} - \boldsymbol{\Theta}^*\|_\infty \cdot \|\mathbf{W}^{(l)}\boldsymbol{\Theta}^*\|_{\infty,\infty} + \|\widehat{\boldsymbol{\Theta}}^{(l)}\widehat{\boldsymbol{\Sigma}}^{(l)} - \mathbf{I}\|_{\infty,\infty} \cdot \|\widehat{\boldsymbol{\Theta}}^{(l)} - \boldsymbol{\Theta}^*\|_\infty. \quad (6.2)$$

For the term $\widehat{\boldsymbol{\Theta}}^{(l)}\widehat{\boldsymbol{\Sigma}}^{(l)} - \mathbf{I}$, we have

$$\|\widehat{\boldsymbol{\Theta}}^{(l)}\widehat{\boldsymbol{\Sigma}}^{(l)} - \mathbf{I}\|_{\infty,\infty} = \|\widehat{\boldsymbol{\Theta}}^{(l)}\widehat{\boldsymbol{\Sigma}}^{(l)} - \widehat{\boldsymbol{\Theta}}^{(l)}\boldsymbol{\Sigma}^* + \widehat{\boldsymbol{\Theta}}^{(l)}\boldsymbol{\Sigma}^* - \mathbf{I}\|_{\infty,\infty}$$
$$= \|\boldsymbol{\Theta}^*(\widehat{\boldsymbol{\Sigma}}^{(l)} - \boldsymbol{\Sigma}^*) + (\widehat{\boldsymbol{\Theta}}^{(l)} - \boldsymbol{\Theta}^*)\boldsymbol{\Sigma}^* + (\widehat{\boldsymbol{\Sigma}}^{(l)} - \boldsymbol{\Sigma}^*)(\widehat{\boldsymbol{\Theta}}^{(l)} - \boldsymbol{\Theta}^*)\|_{\infty,\infty}$$
$$\leq \|\boldsymbol{\Theta}^*\mathbf{W}^{(l)}\|_{\infty,\infty} + \|\widehat{\boldsymbol{\Theta}}^{(l)} - \boldsymbol{\Theta}^*\|_{\infty,\infty} \cdot \|\boldsymbol{\Sigma}^*\|_\infty + \|\widehat{\boldsymbol{\Sigma}}^{(l)} - \boldsymbol{\Sigma}^*\|_{\infty,\infty} \cdot \|\widehat{\boldsymbol{\Theta}}^{(l)} - \boldsymbol{\Theta}^*\|_\infty.$$

Submitting the above inequality into (6.2) yields

$$\|\text{Rem}^{(l)}\|_{\infty,\infty} \leq 2\|\boldsymbol{\Theta}^*\mathbf{W}^{(l)}\|_{\infty,\infty} \cdot \|\widehat{\boldsymbol{\Theta}}^{(l)} - \boldsymbol{\Theta}^*\|_\infty + \|\widehat{\boldsymbol{\Theta}}^{(l)} - \boldsymbol{\Theta}^*\|_{\infty,\infty} \cdot \|\widehat{\boldsymbol{\Theta}}^{(l)} - \boldsymbol{\Theta}^*\|_\infty \cdot \|\boldsymbol{\Sigma}^*\|_\infty$$
$$+ \|\widehat{\boldsymbol{\Sigma}}^{(l)} - \boldsymbol{\Sigma}^*\|_{\infty,\infty} \cdot \|\widehat{\boldsymbol{\Theta}}^{(l)} - \boldsymbol{\Theta}^*\|_\infty^2$$
$$\leq 2\|\boldsymbol{\Theta}^*\|_\infty \cdot \|\mathbf{W}^{(l)}\|_{\infty,\infty} \cdot \|\widehat{\boldsymbol{\Theta}}^{(l)} - \boldsymbol{\Theta}^*\|_\infty + \|\widehat{\boldsymbol{\Theta}}^{(l)} - \boldsymbol{\Theta}^*\|_{\infty,\infty} \cdot \|\widehat{\boldsymbol{\Theta}}^{(l)} - \boldsymbol{\Theta}^*\|_\infty \cdot \|\boldsymbol{\Sigma}^*\|_\infty$$
$$+ \|\widehat{\boldsymbol{\Sigma}}^{(l)} - \boldsymbol{\Sigma}^*\|_{\infty,\infty} \cdot \|\widehat{\boldsymbol{\Theta}}^{(l)} - \boldsymbol{\Theta}^*\|_\infty^2. \quad (6.3)$$

According to the definition of $\mathcal{U}(s,M)$, we have $\|\boldsymbol{\Theta}^*\|_\infty = \|\boldsymbol{\Theta}^{*\top}\|_1 = \|\boldsymbol{\Theta}^*\|_1 \leq M$. Recall that $\mathbf{W}^{(l)} = \widehat{\boldsymbol{\Sigma}}^{(l)} - \boldsymbol{\Sigma}^*$. Applying the upper bound for $\|\widehat{\boldsymbol{\Sigma}}^{(l)} - \boldsymbol{\Sigma}^*\|_{\infty,\infty}$ in Lemma D.6 and the upper bound for $\|\widehat{\boldsymbol{\Theta}}^{(l)} - \boldsymbol{\Theta}^*\|_{\infty,\infty}$ and $\|\widehat{\boldsymbol{\Theta}}^{(l)} - \boldsymbol{\Theta}^*\|_1$ in Lemma D.1 to (6.3), we obtain that

$$\|\text{Rem}^{(l)}\|_{\infty,\infty} \leq 6\pi C_1 M^3 \frac{s\log d}{n} + 4C_0 C_1 M^4 K_{\boldsymbol{\Sigma}^*} \frac{s\log d}{n} + 3\pi C_1^2 M^4 s^2 \left(\frac{\log d}{n}\right)^{3/2} \quad (6.4)$$

holds with probability at least $1 - d^{-1} - d^{-5/2}$. Note that the above inequality holds for any $l = 1, \ldots, m$. Thus we have proved the entries of $\text{Rem}^{(l)}$ are in the order of $O_p(s\log d/n)$, which implies $I_{02}$ is $o_p(1)$ as long as $n > (s\log d)^2$.

**Term $I_{01}$:** Now we are going to show that the entries of $I_{01}$ are asymptotic normal. We use $S_j, S_k$ to denote the support of $\boldsymbol{\Theta}^*_{*j}$ and $\boldsymbol{\Theta}^*_{*k}$. Since $\boldsymbol{\Theta}^* \in \mathcal{U}(s,M)$, we have $|S_j| \leq s, |S_k| \leq s, \forall j,k =$

$1, \cdots, d$. By mean value theorem

$$
\begin{aligned}
\sqrt{n}(\boldsymbol{\Theta}_{*j}^{*\top}\widehat{\boldsymbol{\Sigma}}^{(l)}\boldsymbol{\Theta}_{*k}^{*} - \Theta_{jk}^{*}) &= \sqrt{n}\sum_{p\in S_j, q\in S_k} \Theta_{pj}^{*}\big(\widehat{\Sigma}_{pq}^{(l)} - \Sigma_{pq}^{*}\big)\Theta_{qk}^{*} \\
&= \sqrt{n}\sum_{p\in S_j, q\in S_k} \Theta_{pj}^{*}\Theta_{qk}^{*}\Big(\sin\Big(\widehat{\tau}_{pq}^{(l)}\frac{\pi}{2}\Big) - \sin\Big(\tau_{pq}\frac{\pi}{2}\Big)\Big) \\
&= \sqrt{n}\sum_{p\in S_j, q\in S_k} \Theta_{pj}^{*}\Theta_{qk}^{*}\cos\Big(\tau_{pq}\frac{\pi}{2}\Big)\frac{\pi}{2}\big(\widehat{\tau}_{pq}^{(l)} - \tau_{pq}\big) \\
&\quad - \frac{\sqrt{n}}{2}\sum_{p\in S_j, q\in S_k} \Theta_{pj}^{*}\Theta_{qk}^{*}\sin\Big(\widetilde{\tau}_{pq}\frac{\pi}{2}\Big)\Big(\frac{\pi}{2}(\widehat{\tau}_{pq}^{(l)} - \tau_{pq})\Big)^2,
\end{aligned} \qquad (6.5)
$$

where $\widetilde{\tau}_{pq}$ is a number between $\widehat{\tau}_{pq}$ and $\tau_{pq}$. We first deal with the first term in the sum above. In order to make the notations simpler, we just omit the index $l$ of $\tau_{jk}^{(l)}$ in the next context of this proof. Recall the notations defined in (4.4) and the Hájek's decomposition method in Hoeffding (1948), and we have

$$\widehat{\tau}_{pq} - \tau_{pq} = \frac{2}{n}\sum_{i=1}^{n} h_{pq}^{i} + \frac{2}{n(n-1)}\sum_{1\leq i<i'\leq n} w_{pq}^{ii'}.$$

This gives us the following identity

$$
\begin{aligned}
&\sqrt{n}\sum_{p\in S_j, q\in S_k} \Theta_{pj}^{*}\Theta_{qk}^{*}\cos\Big(\tau_{pq}\frac{\pi}{2}\Big)\frac{\pi}{2}\big(\widehat{\tau}_{pq} - \tau_{pq}\big) \\
&= \underbrace{n^{-\frac{1}{2}}\pi\sum_{i=1}^{n}\sum_{p\in S_j, q\in S_k} \Theta_{pj}^{*}\Theta_{qk}^{*}\cos\Big(\tau_{pq}\frac{\pi}{2}\Big)h_{pq}^{i}}_{I_1} + \underbrace{\frac{\pi}{n^{1/2}(n-1)}\sum_{1\leq i<i'\leq n}\sum_{p\in S_j, q\in S_k} \Theta_{pj}^{*}\Theta_{qk}^{*}\cos\Big(\tau_{pq}\frac{\pi}{2}\Big)w_{pq}^{ii'}}_{I_2}.
\end{aligned}
$$

We show that $I_1$ can be represented as the sum of i.i.d. zero mean terms. By the definition of $\mathbf{M}^i$ in (4.5), we can rewrite $I_1$ as

$$I_1 = n^{-\frac{1}{2}}\sum_{i=1}^{n}\boldsymbol{\Theta}_{*j}^{*\top}\mathbf{M}^{i}\boldsymbol{\Theta}_{*k}^{*}.$$

Let $g_i = \boldsymbol{\Theta}_{*j}^{*\top}\mathbf{M}^{i}\boldsymbol{\Theta}_{*k}^{*}$, by assumption we have

$$\mathrm{Var}(g_i) = \mathbb{E}\Big(\boldsymbol{\Theta}_{*j}^{*\top}\mathbf{M}^{i}\boldsymbol{\Theta}_{*k}^{*}\Big)^2 \geq \rho_{\min}^{i}\|\boldsymbol{\Theta}_{*j}^{*}\|_2^2 \cdot \|\boldsymbol{\Theta}_{*k}^{*}\|_2^2,$$

where constant $\rho_{\min}^{i} > 0$ could be omitted. Denote $s_n^2 := \sum_{i=1}^{n}\mathrm{Var}\big(\boldsymbol{\Theta}_{*j}^{*\top}\mathbf{M}^{i}\boldsymbol{\Theta}_{*k}^{*}\big) = \sum_{i=1}^{n}\mathrm{var}(g_i)$, and then by definition $s_n^2 = n\sigma_{jk,l}^2$. Note that $\sigma_{jk,l}^2$ is indexed by $l$ because $\mathbf{M}^i$ is computed by samples $\boldsymbol{X}_{(l-1)n+1},\ldots,\boldsymbol{X}_{ln}$ for any $(l-1)n+1 \leq i \leq ln$, and $\mathbf{M}^i$ is zero mean. . We need to verify the Lyapunov's condition for CLT:

$$\lim_{n\to\infty} \frac{1}{s_n^{2+\delta}}\sum_{i=1}^{n}\mathbb{E}|g_i|^{2+\delta} = 0.$$

Just set $\delta = 1$ in our case,

$$\frac{1}{s_n^3} \sum_{i=1}^n \mathbb{E}|g_i|^3 \leq \frac{1}{n^{3/2}\rho_{\min}^0 \|\mathbf{\Theta}_{*j}^*\|_2^3 \cdot \|\mathbf{\Theta}_{*k}^*\|_2^3} \sum_{i=1}^n \mathbb{E}|\mathbf{\Theta}_{*j}^{*\top}\mathbf{M}^i\mathbf{\Theta}_{*k}^*|^3 \leq n^{-\frac{1}{2}} \|\mathbf{M}_{S_j,S_k}^i\|_2^3,$$

where we use the fact $|\mathbf{\Theta}_{*j}^{*\top}\mathbf{M}^i\mathbf{\Theta}_{*k}^*| \leq \|\mathbf{\Theta}_{*j}^*\|_2 \cdot \|\mathbf{M}^i\|_2 \cdot \|\mathbf{\Theta}_{*k}^*\|_2$. Since $|M_{pq}^i| \leq 2\pi$, $\|\mathbf{M}_{S_j,S_k}^i\|_2^3 \leq (2\pi)^3$, we have

$$\frac{1}{s_n^3} \sum_{i=1}^n \mathbb{E}|g_i|^3 \leq \frac{(2\pi)^3}{n^{1/2}} = o(1),$$

which means the Lyapunov's condition holds. And furthermore by CLT we have

$$\frac{\sum_{i=1}^n g_i}{s_n} = \frac{n^{-1/2}\sum_{i=1}^n g_i}{\sigma_{jk,l}} = \frac{I_1}{\sigma_{jk,l}} \xrightarrow{d} N(0,1).$$

Next we deal with the second term $I_2$. Note that $I_2$ is biased and we will show that $I_2$ goes to 0 asymptotically. We need to calculate the following first:

$$\mathbb{E}(w_{pq}^{ii'} w_{lr}^{ee'}) = \mathbb{E}\Big(\big(h_{pq}^{ii'} - h_{pq}^{ii'|i} - h_{pq}^{ii'|i'}\big)\big(h_{lr}^{ee'} - h_{lr}^{ee'|e} - h_{lr}^{ee'|e'}\big)\Big)$$

$$= \mathbb{E}(h_{pq}^{ii'} h_{lr}^{ee'}) - \mathbb{E}(h_{pq}^{ii'} h_{lr}^{ee'|e}) - \mathbb{E}(h_{pq}^{ii'} h_{lr}^{ee'|e'}) - \mathbb{E}(h_{pq}^{ii'|i} h_{lr}^{ee'}) + \mathbb{E}(h_{pq}^{ii'|i} h_{lr}^{ee'|e}) + \mathbb{E}(h_{pq}^{ii'|i} h_{lr}^{ee'|e'})$$

$$- \mathbb{E}(h_{pq}^{ii'|i'} h_{lr}^{ee'}) + \mathbb{E}(h_{pq}^{ii'|i'} h_{lr}^{ee'|e}) + \mathbb{E}(h_{pq}^{ii'|i'} h_{lr}^{ee'|e'}).$$

Note that when $p \neq q, l \neq r, i \neq i', e \neq e', e \neq i \neq e', e \neq i' \neq e'$, we have $\mathbb{E}(w_{pq}^{ii'} w_{lr}^{ee'}) = 0$ by independence, since $\mathbb{E}(w_{pq}^{ii'}) = 0, \mathbb{E}(w_{lr}^{ee'}) = 0$. Therefore, when $p \neq q, l \neq r, i \neq i', i \neq e', i' \neq e'$, by similar calculation from above we have

$$\mathbb{E}(w_{pq}^{ii'} w_{lr}^{ie'}) = \mathbb{E}(h_{pq}^{ii'} h_{lr}^{ie'}) - \mathbb{E}(h_{pq}^{ii'} h_{lr}^{ie'|i}) - \mathbb{E}(h_{pq}^{ii'} h_{lr}^{ie'|e'}) - \mathbb{E}(h_{pq}^{ii'|i} h_{lr}^{ie'}) + \mathbb{E}(h_{pq}^{ii'|i} h_{lr}^{ie'|i}) + \mathbb{E}(h_{pq}^{ii'|i} h_{lr}^{ie'|e'})$$

$$- \mathbb{E}(h_{pq}^{ii'|i'} h_{lr}^{ie'}) + \mathbb{E}(h_{pq}^{ii'|i'} h_{lr}^{ie'|i}) + \mathbb{E}(h_{pq}^{ii'|i'} h_{lr}^{ie'|e'})$$

$$= \mathbb{E}(h_{pq}^{ii'} h_{lr}^{ie'}) - \mathbb{E}(h_{pq}^{ii'} h_{lr}^{ie'|i}) - \mathbb{E}(h_{pq}^{ii'|i} h_{lr}^{ie'}) + \mathbb{E}(h_{pq}^{ii'|i} h_{lr}^{ie'|i}), \tag{6.6}$$

where all the other items are 0 by independence. Furthermore, by iterated expectation, we know that all the four remaining terms equals to $\mathbb{E}(h_{pq}^{ii'|i} h_{lr}^{ie'|i})$, which implies $\mathbb{E}(w_{pq}^{ii'} w_{lr}^{ie'}) = 0$.

Note that $\mathbb{E}(w_{pq}^{ii'}) = 0$ also impies $\mathbb{E}(I_2) = 0$, we have

$$\frac{\text{Var}(I_2)}{\sigma_{jk,l}^2} \leq \frac{\mathbb{E}(I_2^2)}{\rho_{\min}^0 \|\mathbf{\Theta}_{*j}^*\|_2^2 \cdot \|\mathbf{\Theta}_{*k}^*\|_2^2}$$

$$= \frac{\pi^2}{\rho_{\min}^0 \|\mathbf{\Theta}_{*j}^*\|_2^2 \cdot \|\mathbf{\Theta}_{*k}^*\|_2^2 n(n-1)^2} \sum_{1 \leq i < i' \leq n} \mathbb{E}\left(\sum_{p \in S_j, q \in S_k} \Theta_{pj}^* \Theta_{qk}^* \cos\left(\tau_{pq}\frac{\pi}{2}\right) w_{pq}^{ii'}\right)^2$$

$$\leq \frac{\pi^2}{\rho_{\min}^0 \|\mathbf{\Theta}_{*j}^*\|_2^2 \cdot \|\mathbf{\Theta}_{*k}^*\|_2^2 n(n-1)^2} \cdot \frac{n(n-1)}{2} \cdot 36\|\mathbf{\Theta}_{*j}^*\|_1^2 \cdot \|\mathbf{\Theta}_{*k}^*\|_1^2$$

$$\leq \frac{18\pi^2 s^2}{\rho_{\min}^0 (n-1)} = o(1),$$

where we used the fact that $|w_{pq}^{ii'}| \leq |h_{pq}^{ii'}| + |h_{pq}^{ii'|i}| + |h_{pq}^{ii'|i'}| \leq 6$ in the second inequality. Recall the Chebyshev's inequality

$$\mathbb{P}\big(|I_2| \geq \sigma_{jk,l}\big) \leq \frac{\text{Var}(I_2)}{\sigma_{jk,l}^2}.$$

Then we have $I_2/\sigma_{jk,l} = o_p(1)$ with probability at least $1 - O(s^2/n)$.

Finally, let's focus on the last term in (6.5). Omit the index $l$ and denote it as $I_3$:

$$I_3 = \frac{\sqrt{n}}{2} \sum_{p \in S_j, q \in S_k} \Theta_{pj}^* \Theta_{qk}^* \sin\left(\widetilde{\tau}_{pq}\frac{\pi}{2}\right)\left(\frac{\pi}{2}(\widehat{\tau}_{pq} - \tau_{pq})\right)^2.$$

Since $\widehat{\tau}_{pq}$ is a U-statistic, and its kernel is a bounded function between $-1$ and $1$ and $\mathbb{E}\widehat{\tau}_{pq} = \tau_{pq}$. Then by Hoeffding's inequality for U-statistics, we obtain

$$\mathbb{P}\left(\sup_{p,q} |\widehat{\tau}_{pq} - \tau_{pq}| > t\right) \leq 2d^2 e^{-\frac{nt^2}{4}}. \tag{6.7}$$

Choosing $t = 4\sqrt{\log d/n}$, we have

$$\frac{I_3}{\sqrt{1/n \sum_{i=1}^n \sigma_{jk,l}^2}} = \frac{1}{\sqrt{1/n \sum_{i=1}^n \sigma_{jk,l}^2}} \cdot \frac{\sqrt{n}}{2} \sum_{p \in S_j, q \in S_k} \Theta_{pj}^* \Theta_{qk}^* \sin\left(\widetilde{\tau}_{pq}\frac{\pi}{2}\right)\left(\frac{\pi}{2}(\widehat{\tau}_{pq} - \tau_{pq})\right)^2$$

$$\leq \frac{\sqrt{n}\pi^2}{8\sqrt{\rho_{\min}^0 \|\Theta_{*j}^*\|_2^2 \cdot \|\Theta_{*k}^*\|_2^2}} \cdot \sqrt{s}\|\Theta_{*j}^*\|_2 \cdot \sqrt{s}\|\Theta_{*k}^*\|_2 \sup_{p,q}(\widehat{\tau}_{pq} - \tau_{pq})^2$$

$$\leq \frac{2\pi^2 s \log d}{\sqrt{\rho_{\min}^0}\sqrt{n}} = O_p\left(\frac{s \log d}{\sqrt{n}}\right) = o_p(1),$$

hold with probability at least $1 - 2/d^2$.

So far, we have proved $Z_{jk}^{(l)}/\sigma_{jk,l} \rightsquigarrow N(0,1)$, $\forall l = 1, \cdots, m$. Thus, by independence of the data on different machine we have

$$Z_{jk} = \frac{1}{m}\sum_{l=1}^m \frac{Z_{jk}^{(l)}}{\sigma_{jk,l}} \rightsquigarrow N\left(0, \frac{1}{m}\right),$$

which completes the proof. $\square$

### 6.4 Proof of Corollary 5.12

Using Theorem 5.9 and Proposition 5.11, we now prove Corollary 5.12.

*Proof of Corollary 5.12.* By Theorem 5.9, we have

$$-\sqrt{\frac{n}{m}} \sum_{l=1}^m \frac{\Theta_{*j}^{*\top}\widehat{\Sigma}^{(l)}\Theta_{*k}^* - \Theta_{jk}^*}{\sigma_{jk}} \rightsquigarrow N(0,1).$$

And by Proposition 5.11, we have $\widehat{\sigma}_{jk}^2 \xrightarrow{p} \sigma_{jk}^2$. Then by Slutsky's Theorem, we have

$$\widehat{U}_n = -\frac{\sqrt{N}}{m}\sum_{l=1}^{m} \frac{\bar{\Theta}_{jk}^{(l)}}{\widehat{\sigma}_{jk}} \rightsquigarrow N(0,1).$$

It follows that $\mathbb{P}(\widehat{U}_n \leq t)$ converges uniformly to the cumulative distribution function of standard normal distribution. $\square$

# 7 Application to Gaussian Graphical Models

The Gaussian graphical model (Meinshausen and Bühlmann, 2006; Yuan and Lin, 2007) is a special case of the transelliptical graphical model we studied in previous sections, where all the monotone univariate functions are chosen to be identity functions. In this section, we show that our proposed distributed algorithm and theory can be applied to Gaussian graphical models straightforwardly.

In the Gaussian graphical model, the data are sampled from a multivariate normal distribution with zero mean and a covariance matrix $\boldsymbol{\Sigma}^*$, i.e., $\boldsymbol{X} = (X_1, \ldots, X_d)^\top \sim N(\mathbf{0}, \boldsymbol{\Sigma}^*)$. The random variables $X_1, \ldots, X_d$ are associated with the vertex set $V = \{1, \ldots, d\}$ of an undirected graph $G = (V, E)$ with an edge set $E$. We say that the precision matrix $\boldsymbol{\Theta}^* = (\boldsymbol{\Sigma}^*)^{-1}$ represents the edge structure of the graph $G$ if $\Theta_{jk}^* = 0$ for all $(j,k) \notin E$. Consequently, estimation of the precision matrix corresponds to parameter estimation in a Gaussian graphical model and specifying the non-zero entries of $\boldsymbol{\Theta}^*$ corresponds to model selection. Note that in the Gaussian case, sparsity assumptions on the entries of the precision matrix translate to sparsity of the edges in the underlying Gaussian graphical model.

In the setting of distributed estimation for Gaussian graphical models, we have data matrices $\mathbf{X}^{(l)} \in \mathbb{R}^{n_l \times d}$, $l = 1, \ldots, m$, distributed on $m$ machines, where $n_l$ is the number of observations on the $l$-th machine. Each row of $\mathbf{X}^{(l)}$ is sampled independently from $N(\mathbf{0}, \boldsymbol{\Sigma}^*)$. Our goal is to estimate the precision matrix $\boldsymbol{\Theta}^*$ based on $\mathbf{X}^{(l)}$, $l = 1, \ldots, m$. Without loss of generality, we assume that $n_1 = \ldots = n_m = n$ and total sample size $N = nm$.

## 7.1 Distributed Estimation for Gaussian Graphical Models

Recall the estimation procedure presented in Section 4.2, our distributed algorithm for Gaussian graphical models consists of four steps: (1) On each machine, we estimate the latent precision matrix $\widehat{\boldsymbol{\Theta}}^{(l)}$ based on $\mathbf{X}^{(l)}$ using the rank-based estimator in Section 3.2, which is sparse; (2) For each machine, we debias the estimator $\widehat{\boldsymbol{\Theta}}^{(l)}$ as $\widetilde{\boldsymbol{\Theta}}^{(l)} = 2\widehat{\boldsymbol{\Theta}}^{(l)} - \widehat{\boldsymbol{\Theta}}^{(l)}\widehat{\boldsymbol{\Sigma}}^{(l)}\widehat{\boldsymbol{\Theta}}^{(l)}$, where $\widehat{\boldsymbol{\Sigma}}^{(l)} \in \mathbb{R}^{d \times d}$ is the estimator of correlation matrix. Note that $\widetilde{\boldsymbol{\Theta}}^{(l)}$ is no longer sparse; (3) We send the debiased estimators $\widetilde{\boldsymbol{\Theta}}^{(l)}$, $l = 1, \ldots, m$ to a central machine and average them as $\bar{\boldsymbol{\Theta}} = 1/m \sum_{l=1}^{m} \widetilde{\boldsymbol{\Theta}}^{(l)}$, which is still dense; (4) We threshold the averaged debiased estimator $\bar{\boldsymbol{\Theta}}$ to obtain a sparse precision matrix estimator, i.e., $\check{\boldsymbol{\Theta}} = \text{HT}(\bar{\boldsymbol{\Theta}}, t)$, where $\text{HT}(\cdot, t)$ is defined in (4.2) and $t$ is a threshold that depends on $m, n$ and $d$.

Now we are ready to present our main theory for the distributed estimation for Gaussian graphical models.

**Theorem 7.1.** Under Assumptions 5.1 and 5.2, if the thresholding parameter is chosen to be $t = M^2 K_{\boldsymbol{\Sigma}^*}\sqrt{\log d/N} + C_1 M^4 K_{\boldsymbol{\Sigma}^*} s \log d/n$, then with probability at least $1 - 2/d - 1/d^{C_2-2}$, the estimator $\check{\boldsymbol{\Theta}}$ satisfies

$$\|\check{\boldsymbol{\Theta}} - \boldsymbol{\Theta}^*\|_{\infty,\infty} \leq 2M^2 K_{\boldsymbol{\Sigma}^*}\sqrt{\frac{\log d}{N}} + 2C_1 M^4 K_{\boldsymbol{\Sigma}^*}\frac{s\log d}{n},$$

$$\|\check{\boldsymbol{\Theta}} - \boldsymbol{\Theta}^*\|_2 \leq 2M^2 K_{\boldsymbol{\Sigma}^*} s\sqrt{\frac{\log d}{N}} + 2C_1 M^4 K_{\boldsymbol{\Sigma}^*}\frac{s^2\log d}{n},$$

$$\|\check{\boldsymbol{\Theta}} - \boldsymbol{\Theta}^*\|_F \leq 2M^2 K_{\boldsymbol{\Sigma}^*}\sqrt{\frac{sd\log d}{N}} + 2C_1 M^4 K_{\boldsymbol{\Sigma}^*}\frac{s^{3/2}\sqrt{d}\log d}{n},$$

where $C_1, C_2 > 0$ are absolute constants.

**Remark 7.2.** Though Theorem 7.1 does not require the specific choice of $m$, if we choose $m \lesssim \sqrt{N/(s^2 \log d)}$, then the distributed estimator attains the same rate of convergence as the centralized estimator. In addition, when $m$ is chosen properly such that $m \lesssim \sqrt{N/(s^2 \log d)}$, the distributed estimator can achieve model selection consistency under the condition that $\min_{(j,k)\in\mathbb{R}^{d\times d}} \Theta^*_{jk} \geq CM^2\sqrt{\log d/N}$. This is comparable with the conditions for model selection consistency in CLIME method (Cai et al., 2011) and graphical Lasso (Ravikumar et al., 2011). In a recent independent work, Arroyo and Hou (2016) proposed a distributed estimation method for Gaussian graphical models, which has an identical scaling condition on number of machines. However, their theoretical guarantee requires the irrepresentable condition (Ravikumar et al., 2011; Jankova and van de Geer, 2013), which is very stringent. In sharp contrast, our theory does not require such a condition.

### 7.2 Distributed Asymptotic Inference for Gaussian Graphical Models

Similar to the transelliptical graphical model, we establish the asymptotic normality for each entry in the distributed estimator of precision matrix.

**Theorem 7.3.** Suppose that $\boldsymbol{X}_1, \cdots, \boldsymbol{X}_N$ are independently and identically distributed with $\mathbb{E}\boldsymbol{X}_1 = \boldsymbol{0}, \mathrm{cov}(\boldsymbol{X}_1) = \boldsymbol{\Sigma}^*$. Let $\boldsymbol{\Theta}^* = (\boldsymbol{\Sigma}^*)^{-1}$, and assume that Assumptions 5.1 and 5.2 hold, and $s\log d/\sqrt{n} = o(1)$, where $n = N/m$ is the sample size on each machine and $m$ is the number of machines. For any $j, k = 1, \cdots, d$ and $i = 1, \ldots, n$, we denote $\sigma^2_{jk} := \mathrm{Var}(1/m \sum_{l=1}^m \boldsymbol{\Theta}^*_{j*}\boldsymbol{X}_{(l-1)n+i}\boldsymbol{\Theta}^*_{k*}\boldsymbol{X}_{ln})$. Then

$$\sigma^2_{jk} = 1/m\big(\Theta^*_{jj}\Theta^*_{kk} + \Theta^{*2}_{jk}\big),$$

which is finite. Therefore, we have

$$\frac{\sqrt{n}(\bar{\Theta}_{jk} - \Theta^*_{jk})}{\sigma_{jk}} = -\frac{\sqrt{n}}{m}\sum_{l=1}^m \frac{\boldsymbol{\Theta}^*_{j*}\widehat{\boldsymbol{\Sigma}}^{(l)}\boldsymbol{\Theta}^*_{*k} - \Theta^*_{jk}}{\sigma_{jk}} + o_p(1) = Z^n_{jk} + o_p(1),$$

where $Z^n_{jk}$ converges weakly to $N(0, 1)$. Furthermore, since we have $\widehat{\boldsymbol{\Theta}}^{(l)} \xrightarrow{p} \boldsymbol{\Theta}^*$, by Slutsky's theorem, we obtain

$$-\frac{\sqrt{N}}{m}\sum_{l=1}^m \frac{\bar{\Theta}_{jk} - \Theta^*_{jk}}{\sqrt{\widehat{\Theta}^{(l)}_{jj}\widehat{\Theta}^{(l)}_{kk} + (\widehat{\Theta}^{(l)}_{jk})^2}} \rightsquigarrow N(0, 1).$$

When $m = 1$, our method reduces to the inference for centralized Gaussian graphical models. And our results are consistent with the asymptotic properties for centralized graphical Lasso estimator proved in Jankova and van de Geer (2013). Denote $\widehat{\sigma}_{jk,l}^2 := 1/m\big(\widehat{\Theta}_{jj}^{(l)}\widehat{\Theta}_{kk}^{(l)} + \big(\widehat{\Theta}_{jk}^{(l)}\big)^2\big)$, then we have $\widehat{\sigma}_{jk,l}^2 \xrightarrow{p} \sigma_{jk}^2$ by Slutsky's theorem and the fact that $\widehat{\Theta}^{(l)} \xrightarrow{p} \Theta^*$. By Theorem 7.3, we have $\sqrt{n}/m \sum_{l=1}^{m} (\bar{\Theta}_{jk} - \Theta_{jk}^*)/\widehat{\sigma}_{jk,l} \rightsquigarrow N(0, 1)$, and it follows that the $1 - \alpha$ asymptotic confidence interval for $\Theta_{jk}^*$ is given by

$$\left[\bar{\Theta}_{jk} - \frac{mu_{1-\alpha/2}}{\sqrt{n}}\left(\sum_{l=1}^{m}\frac{1}{\widehat{\sigma}_{jk,l}}\right)^{-1}, \bar{\Theta}_{jk} + \frac{mu_{1-\alpha/2}}{\sqrt{n}}\left(\sum_{l=1}^{m}\frac{1}{\widehat{\sigma}_{jk,l}}\right)^{-1}\right],$$

where $\widehat{\sigma}_{jk,l}^2 := 1/m\big(\widehat{\Theta}_{jj}^{(l)}\widehat{\Theta}_{kk}^{(l)} + \big(\widehat{\Theta}_{jk}^{(l)}\big)^2\big)$ and $u_{1-\alpha/2}$ is the $1 - \alpha/2$ quantile of standard normal distribution. Under $H_0 : \Theta_{jk}^* = 0$, the test statistic can be written as

$$\widehat{U}_n = -\frac{\sqrt{N}}{m}\sum_{l=1}^{m}\frac{\bar{\Theta}_{jk}}{\sqrt{\widehat{\Theta}_{jj}^{(l)}\widehat{\Theta}_{kk}^{(l)} + \big(\widehat{\Theta}_{jk}^{(l)}\big)^2}}.$$

By Theorem 7.3, the asymptotic distribution of $\widehat{U}_n$ should follow standard normal distribution. Then we have the following corollary:

**Corollary 7.4.** Under the same assumptions as Theorem 7.3, under $H_0 : \Theta_{jk}^* = 0$, we have

$$\lim_{n\to\infty}\sup_{t\in\mathbb{R}}\left|\mathbb{P}\big(\widehat{U}_n \leq t\big) - \Phi(t)\right| = 0, \tag{7.1}$$

where $\Phi(\cdot)$ is the cdf of standard normal distribution.

Under significance level $\alpha = 0.05$, the null hypothesis is rejected if $|\widehat{U}_n| > u_{1-\alpha/2}$, and the associated $p$-value is given by $p = 2\big(1 - \Phi\big(|\widehat{U}_n|\big)\big)$.

# 8 Experiments

In this section, we conduct experiments on synthetic data to verify the performance of our proposed distributed algorithm. We first present the results for the point estimation of the latent precision matrix, then we present the simulation results for asymptotic inference.

## 8.1 Simulations for Estimation

In this experiment, we compare the performance of our proposed distributed estimator for transelliptical graphical models (**DistTGM**) on precision matrix estimation with centralized estimator (**Centralized**), the debiased centralized estimator (**Debiased**), and the naive averaged distributed estimator (**NaiveDist**). The centralized estimator makes use of all the samples on one machine and is estimated by Cai et al. (2011). In order to investigate the performance of the debiasing process, we also compare with the debiased centralized estimator, which employs all the samples and performs the debiasing process in (4.1) and the threshold process in (4.2). The naive averaged distributed estimator is given by $\bar{\boldsymbol{\Theta}} = 1/m \sum_{l=1}^{m} \widehat{\boldsymbol{\Theta}}^{(l)}$, where $\widehat{\boldsymbol{\Theta}}^{(l)}$ is the estimated precision matrix based on samples on the $l$-th machine. Unlike our proposed distributed estimator in Section 4.2, the naive averaged estimator is not debiased.

We conducted experiments under the following two settings: (1) to investigate the effect of varying number of machines, we set the total sample size $N = 10,000$ as fixed, and then changed $m$ as well as the sample size $n = N/m$ on each machine; (2) to investigate the effect of varying the total sample size, we fixed the sample size on each machine $n = 100$ and changed the number of machines $m$ as well as the total sample size $N = n \times m$. In both settings, the scaling of $m$ satisfies the requirement in Corollary 5.4.

For the data generating, we considered data from nonparanormal and transelliptical distributions. Specifically, we first generated the inverse covariance $(\boldsymbol{\Sigma}^*)^{-1}$ using **huge** package [1], with dimension size $d = 200$. For the nonparanormal graphical model, we considered the following generating scheme: $\boldsymbol{X} = (x_1, \ldots, x_d)^\top \sim NPN(\boldsymbol{\Sigma}^*; f_1, \ldots, f_d)$, with $x_1 = f_1^{-1}(y_1), \ldots, x_d = f_d^{-1}(y_d)$ and $f_1^{-1}(\cdot) = \ldots = f_d^{-1}(\cdot) = \text{sign}(\cdot) \cdot |^{1/2}/\sqrt{\int |t|\phi(t)dt}$, where $\boldsymbol{Y} = (y_1, \ldots, y_d)^\top$ was generated from Gaussian distribution $N(\boldsymbol{0}, \boldsymbol{\Sigma}^*)$. For the transelliptical graphical model, we considered the following generating scheme: $\boldsymbol{X} \sim \text{TE}_d(\boldsymbol{\Sigma}^*, \xi; f_1, \ldots, f_d)$, where $\xi \sim \chi_d$ and $f_1^{-1}(\cdot) = \ldots = f_d^{-1}(\cdot) = \text{sign}(\cdot) \cdot |\cdot|^3$.

We use the following metrics to characterize the performances of algorithms involved in comparison: the Frobenius, spectral and $\ell_{\infty,\infty}$ norms of precision matrix estimation errors. Additionally, to measure the support recovery, $F_1$ score is introduced to measure the overlap of estimated supports and true supports, which is defined as

$$F_1 = \frac{2 \cdot \text{precision} \cdot \text{recall}}{\text{precision} + \text{recall}},$$

where $\text{precision} = |\text{supp}(\check{\boldsymbol{\Theta}}) \cap \text{supp}(\boldsymbol{\Theta}^*)|/|\text{supp}(\check{\boldsymbol{\Theta}})|$, $\text{recall} = |\text{supp}(\check{\boldsymbol{\Theta}}) \cap \text{supp}(\boldsymbol{\Theta}^*)|/|\text{supp}(\boldsymbol{\Theta}^*)|$, $\check{\boldsymbol{\Theta}}$ is the estimator outputted by Algorithm 1, and $|\cdot|$ is the cardinality of a set. For the parameter tuning, we used grid search to find the best regularization parameters $\lambda$'s in (3.3) for all the methods, and the threshold parameters $t$'s in (4.2) for **DistTGM** and **Debiased**.

Figure 1 shows how the $F_1$ score and the estimation error (in terms of matrix Frobenius, spectral and $\ell_{\infty,\infty}$ norms) change with the varying number of machines $m$ while total sample size $N$ fixed, i.e., under Setting (1). For all methods we report the average and standard error of $F_1$ score and estimation errors over 10 repetitions. In most cases, the two centralized methods tend to output better estimators in terms of support discovery and estimation for precision matrix. It can be seen from Figure 1 that the performance of the proposed distributed estimator **DistTGM** is comparable with the centralized estimators when $m$ is small, while the naive distributed estimator performs much worse than the other three. We also report the experiment time in Table 2 for all the methods except **Debiased**, since it definitely costs more time than the **Centralized** estimator due to the extra debiasing process. In Table 2, when $m$ grows, the time consumption of the centralized estimator keeps the same while those of distributed estimators decreases rapidly in both nonparanormal and transelliptical cases. It is also worth noting that the computational time of the proposed **DistTGM** is slightly higher than the naive distributed estimator, implying that the debiasing step is not the major computational overhead. Overall, **DistTGM** runs much faster than the centralized estimator while it can also attain the same rate of convergence rate of estimation errors when $m$ is large.

---
[1] Available on http://cran.r-project.org/web/packages/huge

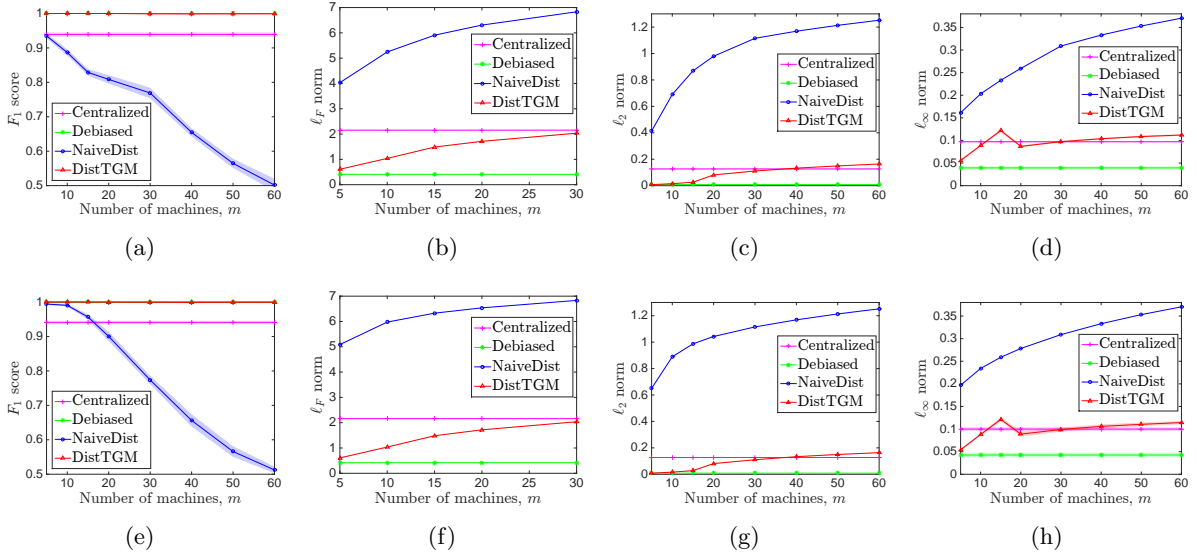

Figure 1: **Setting (1)**: the total sample size $N$ is fixed as $10,000$. Figures 1(a)-1(d) show the $F_1$ score and estimation errors (in terms of Frobenius, spectral and $\ell_{\infty,\infty}$ norms) of different methods for nonparanormal data. Figures 1(e)-1(h) show the $F_1$ score and estimation errors (in terms of Frobenius, spectral and $\ell_{\infty,\infty}$ norms) of different methods for transelliptical data.

Figure 2 shows how the $F_1$ score and the estimation error (in terms of matrix Frobenius, spectral, and $\ell_{\infty,\infty}$ norms) change with the varying number of machines $m$ while sample size on each machine $n = 100$ is fixed, i.e., under Setting (2). It can be seen from Figure 2 that the proposed distributed estimator **DistTGM** is comparable with the centralized estimators in both support discovery and estimation for precision matrix when $m$ is small. Still the naive distributed estimator performs much worse than the other three. Since the total sample size $N$ is increasing when $m$ increases, the estimation performances of both the centralized and the distributed algorithms are improving as $m$ goes up. When number of machines $m$ is small, the **DistTGM** estimator performs slightly better than the **Centralized** estimator due to the debiasing process, which is consistent with the results in Setting (1). Time comparison is presented in Table 3 as well. For the centralized estimator, the total CPU time is reported. For two distributed estimators, we report the CPU time consumed in one machine. It can be seen from Table 3 that the time consumption of the centralized estimator grows linearly as $m$, however the naive distributed estimator and the proposed estimator do not consume more time when $m$ grows. This is because in Setting (2), the sample size in each machine does not change with $m$, while the time consumption of distributed methods only depend on the sample size in each machine. Similar with Table 2, **DistTGM** costs comparable time with **NaiveDist** for each fixed $m$. This again reveals that the debiasing step consumes little time.

Table 2: **Setting** (1): Time (in second) comparison for our proposed distributed estimator versus centralized estimator and naive distributed estimator. In this setting, the total sample size $N$ is fixed as $10,000$.

| | Model Type | Nonparanormal | | | Transelliptical | | |
|---|---|---|---|---|---|---|---|
| | Method | Centralized | NaiveDist | DistTGM | Centralized | NaiveDist | DistTGM |
| No. of Machines | $m=5$ | 18.932±0.231 | 3.372±0.049 | 3.539±0.057 | 18.732±0.275 | 3.324±0.044 | 3.552±0.012 |
| | $m=10$ | 18.932±0.231 | 1.677±0.011 | 1.874±0.008 | 18.732±0.275 | 1.566±0.005 | 1.873±0.006 |
| | $m=15$ | 18.932±0.231 | 1.113±0.006 | 1.403±0.006 | 18.732±0.275 | 1.096±0.008 | 1.402±0.006 |
| | $m=20$ | 18.932±0.231 | 0.813±0.003 | 0.942±0.003 | 18.732±0.275 | 0.813±0.005 | 0.944±0.005 |
| | $m=30$ | 18.932±0.231 | 0.599±0.002 | 0.727±0.002 | 18.732±0.275 | 0.599±0.002 | 0.726±0.002 |
| | $m=40$ | 18.932±0.231 | 0.457±0.001 | 0.584±0.001 | 18.732±0.275 | 0.458±0.002 | 0.584±0.001 |
| | $m=50$ | 18.932±0.231 | 0.406±0.002 | 0.533±0.001 | 18.732±0.275 | 0.407±0.001 | 0.532±0.001 |
| | $m=60$ | 18.932±0.231 | 0.364±0.001 | 0.491±0.001 | 18.732±0.275 | 0.365±0.001 | 0.491±0.001 |

Table 3: **Setting** (2): Time (in second) comparison for our proposed distributed estimator versus centralized estimator and naive distributed estimator. In this setting, the sample size on each machine $n$ is fixed as 100.

| | Model Type | Nonparanormal | | | Transelliptical | | |
|---|---|---|---|---|---|---|---|
| | Method | Centralized | NaiveDist | DistTGM | Centralized | NaiveDist | DistTGM |
| No. of Machines | $m=5$ | 0.905±0.261 | 0.507±0.012 | 0.512±0.009 | 1.021±0.102 | 0.489±0.011 | 0.504±0.012 |
| | $m=10$ | 1.714±0.857 | 0.515±0.029 | 0.526±0.024 | 1.742±0.012 | 0.486±0.011 | 0.496±0.003 |
| | $m=15$ | 2.608±0.185 | 0.514±0.009 | 0.520±0.023 | 2.610±0.032 | 0.484±0.003 | 0.496±0.003 |
| | $m=20$ | 3.377±0.299 | 0.517±0.009 | 0.539±0.021 | 3.409±0.042 | 0.483±0.003 | 0.497±0.003 |
| | $m=30$ | 5.245±0.396 | 0.511±0.008 | 0.517±0.014 | 5.295±0.062 | 0.483±0.001 | 0.496±0.002 |

## 8.2 Simulations for Hypothesis Testing

In this experiment, we illustrate the theoretical results for inference on simulated data and demonstrate the performance of the proposed algorithm for transelliptical graphical models and nonparanormal graphical models. We fix the number of total samples as $N = 2,000$ and dimension $d = 200$.

We examine our inference results for the hypothesis test with null hypothesis $H_0 : \Theta^*_{jk} = 0$, using the test statistic $\widehat{U}_n$ defined in (4.8). Data were generated in the same way as the aforementioned experiments for estimation. In particular, we set the ground-truth $\Theta^*_{jk} = \mu \in [0, 1]$. When $\mu = 0$, we report the type I errors on 500 repetitions. We use the empirical rejection rate to evaluate type I error. Table 4 summarizes the type I errors with significance level $\alpha = 0.05$. We can see that our method achieves accurate type I error and is comparable to the centralized method.

Next we compare the power of hypothesis test $H_0 : \Theta^*_{jk} = \mu$ at 0.05 significance level. When $\mu$ varies between 0 and 1, we report power over 500 repetitions. We use the empirical rejection rate to evaluate power. For each $\mu \in (0, 1]$, we chose the regularization parameter $\lambda$ that was tuned by grid search to maximize power while the corresponding type I error for $\mu = 0$ was controlled under 0.05. Figure 3 shows the power curves of the hypothesis tests with number of machines $m$ varying. We can see that in both nonparanormal and transelliptical settings, the power of our distributed test method is comparable to the centralized test method, when $m$ is relatively small, such as $m = 5$ or 10. In addition, from Table 4 and Figure 3, we can further conclude that the performance of the distributed inference method degrades as the numbers of machine $m$ increases.

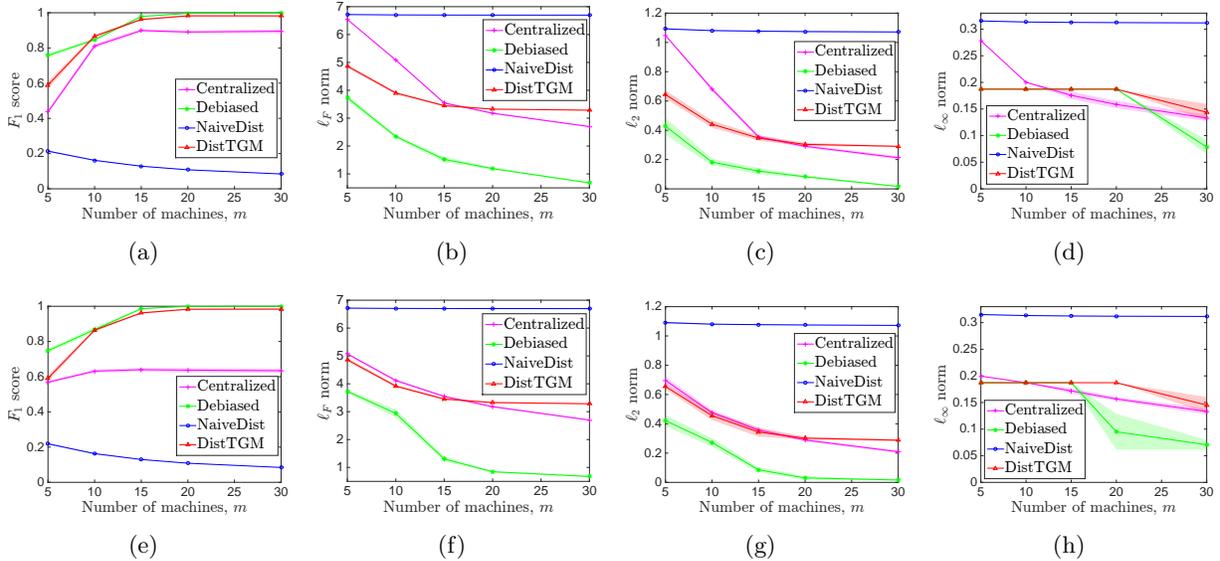

Figure 2: **Setting** (2): the sample size $n$ on each machine is fixed as 100. Figures 2(a)-2(d) show the $F_1$ score and estimation errors (in terms of Frobenius, spectral and $\ell_{\infty,\infty}$ norms) of different methods for nonparanormal data. Figures 2(e)-2(h) show the $F_1$ score and estimation errors (in terms of Frobenius, spectral and $\ell_{\infty,\infty}$ norms) of different methods for transelliptical data.

Table 4: Type I errors for testing $H_0 : \Theta_{jk}^* = 0$ with significance level $\alpha = 0.05$. We run 500 experiments to count the empirical rejection rate for both nonparanormal and transelliptical data under different number of machines.

| | Number of Machines | | | | | |
|---|---|---|---|---|---|---|
| Model Type | $m = 1$(Centralized) | $m = 5$ | $m = 10$ | $m = 15$ | $m = 20$ | $m = 30$ |
| Nonparanormal | 0.052 | 0.048 | 0.062 | 0.058 | 0.056 | 0.046 |
| Transelliptical | 0.044 | 0.058 | 0.052 | 0.064 | 0.042 | 0.054 |

## 9 Conclusions

In this paper, we propose communication-efficient distributed estimation and inference methods for transelliptical graphical models in the high dimensional regime. The key point in our method is obtaining a debiased estimator for latent precision matrix on each machine and then aggregating them on a central machine. We theoretically analyze the proper scaling of $m$ in order to attain the same statistical performance as the centralized estimator and centralized test statistic. In addition, we discuss the application of our distributed method to Gaussian graphical models, which are special cases of ours. Numerical simulations for both estimation and inference support our theory.

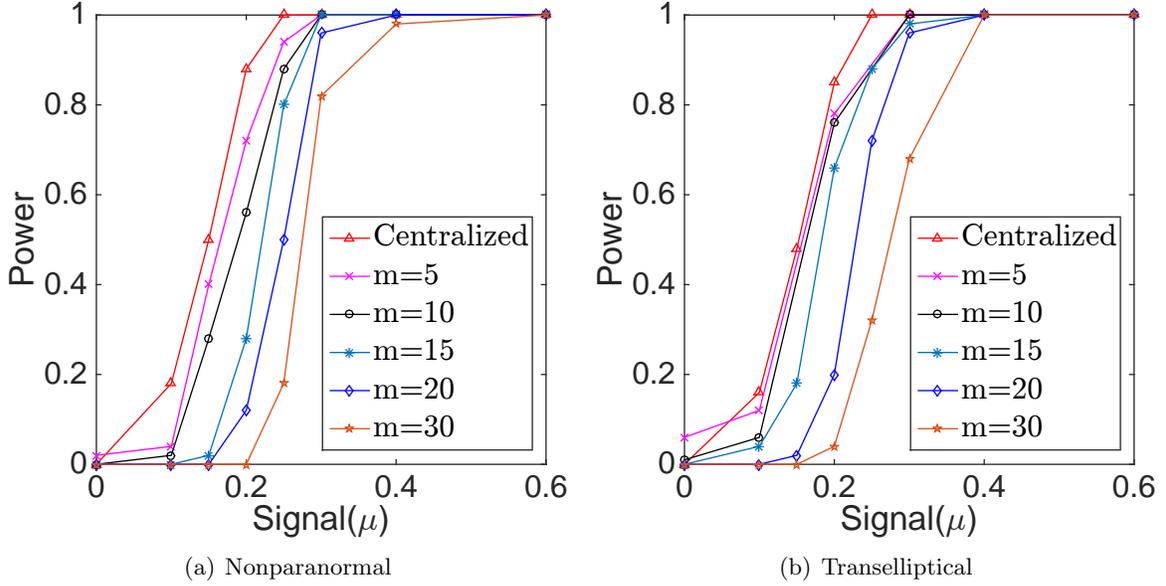

(a) Nonparanormal   (b) Transelliptical

Figure 3: Power curves for testing $H_0$: $\Theta^*_{jk} = 0$ with significance level $\alpha = 0.05$. We run 500 experiments to count the empirical rejection rate for both nonparanormal and transelliptical data when the number of machines varies.

## A   Proofs of Technical Lemmas for Main Theory

In this section, we prove the lemmas we used in Sections 5 and 6. First, we are going to prove Proposition 5.11, which is useful in our main theory for hypothesis testing analysis.

*Proof of Proposition 5.11.* Recall the definition of $\mathbf{M}^i$ in (4.5), where $M^i_{pq} = \pi \cos\left(\tau_{pq} \frac{\pi}{2}\right) h^i_{pq}$. Note that $\mathbb{E}(\mathbf{M}^i) = \mathbf{0}$ due to $\mathbb{E} h^i_{pq} = 0$, and thus $\mathrm{Var}(\boldsymbol{\Theta}^{*\top}_{*j} \mathbf{M}^i \boldsymbol{\Theta}^*_{*k}) = \mathbb{E}(\boldsymbol{\Theta}^{*\top}_{*j} \mathbf{M}^i \boldsymbol{\Theta}^*_{*k})^2$. Then we have

$$\left| \frac{1}{n} \sum_{i=1}^n \big(\widehat{\boldsymbol{\Theta}}^\top_{*j} \widehat{\mathbf{M}}^i \widehat{\boldsymbol{\Theta}}_{*k}\big)^2 - \mathbb{E}(\boldsymbol{\Theta}^{*\top}_{*j} \mathbf{M}^i \boldsymbol{\Theta}^*_{*k})^2 \right| \leq \underbrace{\left| \frac{1}{n} \sum_{i=1}^n \big[\big(\widehat{\boldsymbol{\Theta}}^\top_{*j} \widehat{\mathbf{M}}^i \widehat{\boldsymbol{\Theta}}_{*k}\big)^2 - \big(\boldsymbol{\Theta}^{*\top}_{*j} \widehat{\mathbf{M}}^i \boldsymbol{\Theta}^*_{*k}\big)^2\big] \right|}_{I_1}$$

$$+ \underbrace{\left| \frac{1}{n} \sum_{i=1}^n \big(\boldsymbol{\Theta}^{*\top}_{*j} \widehat{\mathbf{M}}^i \boldsymbol{\Theta}^*_{*k}\big)^2 - \mathbb{E}(\boldsymbol{\Theta}^{*\top}_{*j} \mathbf{M}^i \boldsymbol{\Theta}^*_{*k})^2 \right|}_{I_2}.$$

For term $I_1$, we further have

$$I_1 \leq \underbrace{\left|\frac{1}{n}\sum_{i=1}^{n}\left[(\widehat{\boldsymbol{\Theta}}_{*j}^{\top}\widehat{\mathbf{M}}^i\widehat{\boldsymbol{\Theta}}_{*k})^2 - (\boldsymbol{\Theta}_{*j}^{*\top}\widehat{\mathbf{M}}^i\widehat{\boldsymbol{\Theta}}_{*k})^2\right]\right|}_{I_{11}} + \underbrace{\left|\frac{1}{n}\sum_{i=1}^{n}\left[(\boldsymbol{\Theta}_{*j}^{*\top}\widehat{\mathbf{M}}^i\widehat{\boldsymbol{\Theta}}_{*k})^2 - (\boldsymbol{\Theta}_{*j}^{*\top}\widehat{\mathbf{M}}^i\boldsymbol{\Theta}_{*k}^*)^2\right]\right|}_{I_{12}}.$$

First, for term $I_{11}$ we have

$$\begin{aligned}
I_{11} &= \left|\frac{1}{n}\sum_{i=1}^{n}(\widehat{\boldsymbol{\Theta}}_{*j} - \boldsymbol{\Theta}_{*j}^*)^{\top}\widehat{\mathbf{M}}^i\widehat{\boldsymbol{\Theta}}_{*k}\widehat{\boldsymbol{\Theta}}_{*k}^{\top}\widehat{\mathbf{M}}^i(\widehat{\boldsymbol{\Theta}}_{*j} + \boldsymbol{\Theta}_{*j}^*)\right| \\
&\leq \frac{1}{n}\sum_{i=1}^{n}\left|(\widehat{\boldsymbol{\Theta}}_{*j} - \boldsymbol{\Theta}_{*j}^*)^{\top}\widehat{\mathbf{M}}^i\widehat{\boldsymbol{\Theta}}_{*k}\right| \cdot \left|\widehat{\boldsymbol{\Theta}}_{*k}^{\top}\widehat{\mathbf{M}}^i(\widehat{\boldsymbol{\Theta}}_{*j} + \boldsymbol{\Theta}_{*j}^*)\right| \\
&\leq \frac{1}{n}\sum_{i=1}^{n}\|\widehat{\boldsymbol{\Theta}}_{*j} - \boldsymbol{\Theta}_{*j}^*\|_1 \cdot \|\widehat{\mathbf{M}}^i\|_{\infty,\infty} \cdot \|\widehat{\boldsymbol{\Theta}}_{*k}\|_1 \cdot \|\widehat{\boldsymbol{\Theta}}_{*k}\|_1 \cdot \|\widehat{\mathbf{M}}^i\|_{\infty,\infty} \cdot \|\widehat{\boldsymbol{\Theta}}_{*j} - \boldsymbol{\Theta}_{*j}^*\|_1,
\end{aligned}$$

where the first inequality is due to the triangle's inequality, and the last inequality is due to Hölder's inequality and the fact that $\|\mathbf{Ax}\|_1 \leq \|\mathbf{A}\|_{\infty,\infty} \cdot \|\mathbf{x}\|_1$ for a matrix $\mathbf{A}$ and a vector $\mathbf{x}$. Thus we obtain

$$I_{11} \leq \|\widehat{\boldsymbol{\Theta}}_{*j} - \boldsymbol{\Theta}_{*j}^*\|_1 \cdot \|\widehat{\boldsymbol{\Theta}}_{*j} + \boldsymbol{\Theta}_{*j}^*\|_1 \cdot \|\widehat{\boldsymbol{\Theta}}_{*k}\|_1^2 \cdot (2\pi)^2,$$

where we use the fact that $\|\widehat{\mathbf{M}}^i\|_{\infty,\infty} \leq 2\pi$ by its definition in (4.7). And by Lemma D.1 we have $\|\widehat{\boldsymbol{\Theta}}_{*j} - \boldsymbol{\Theta}_{*j}^*\|_1 \leq \|\widehat{\boldsymbol{\Theta}} - \boldsymbol{\Theta}^*\|_1 = O_p(s\sqrt{\log d/n})$ for any $j = 1,\ldots,d$.

$$\begin{aligned}
I_{11} &\leq \left(\|\widehat{\boldsymbol{\Theta}}_{*j} - \boldsymbol{\Theta}_{*j}^*\|_1^2 + 2\|\widehat{\boldsymbol{\Theta}}_{*j} - \boldsymbol{\Theta}_{*j}^*\|_1 \cdot \|\boldsymbol{\Theta}_{*j}^*\|_1\right)\|\boldsymbol{\Theta}_{*k}^*\|_1^2 \cdot (2\pi)^2 \\
&= O_p(\|\boldsymbol{\Theta}_{*j}^*\|_1^2 \cdot \|\boldsymbol{\Theta}_{*k}^*\|_1^2 s\sqrt{\log d/n}) \\
&= O_p(M^4 s\sqrt{\log d/n}) = o_p(1),
\end{aligned}$$

where the last equality is due to assumption (5.7). Similarly, we obtain

$$I_{12} = O_p(\|\boldsymbol{\Theta}_{*j}^*\|_1^2 \cdot \|\boldsymbol{\Theta}_{*j}^*\|_1^2 s\sqrt{\log d/n}) = o_p(1).$$

For term $I_2$, triangle inequality yields

$$I_2 \leq \underbrace{\left|\frac{1}{n}\sum_{i=1}^{n}\left[(\boldsymbol{\Theta}_{*j}^{*\top}\widehat{\mathbf{M}}^i\boldsymbol{\Theta}_{*k}^*)^2 - (\boldsymbol{\Theta}_{*j}^{*\top}\mathbf{M}^i\boldsymbol{\Theta}_{*k}^*)^2\right]\right|}_{I_{21}} + \underbrace{\left|\frac{1}{n}\sum_{i=1}^{n}(\boldsymbol{\Theta}_{*j}^{*\top}\mathbf{M}^i\boldsymbol{\Theta}_{*k}^*)^2 - \mathbb{E}(\boldsymbol{\Theta}_{*j}^{*\top}\mathbf{M}^i\boldsymbol{\Theta}_{*k}^*)^2\right|}_{I_{22}}. \tag{A.1}$$

We first bound term $I_{21}$. Recall that $\|\mathbf{M}^i\|_{\infty,\infty} \leq 2\pi$, so we have

$$\begin{aligned}
I_{21} &= \left|\frac{1}{n}\sum_{i=1}^{n}\boldsymbol{\Theta}_{*j}^{*\top}(\widehat{\mathbf{M}}^i - \mathbf{M}^i)\boldsymbol{\Theta}_{*k}^*\boldsymbol{\Theta}_{*k}^{*\top}(\widehat{\mathbf{M}}^i + \mathbf{M}^i)\boldsymbol{\Theta}_{*j}^*\right| \\
&\leq \|\boldsymbol{\Theta}_{*j}^*\|_1^2 \cdot \|\boldsymbol{\Theta}_{*k}^*\|_1^2 \cdot (2\pi + 2\pi)\max_{i=1,\ldots,n}\|\widehat{\mathbf{M}}^i - \mathbf{M}^i\|_{\infty,\infty}.
\end{aligned}$$

We need to bound $\max_{i=1,\ldots,n}\|\widehat{\mathbf{M}}^i - \mathbf{M}^i\|_{\infty,\infty}$. Note that $\cos(\cdot)$ is Lipschitz with constant 1, i.e., $\left|\cos(\widehat{\tau}_{pq}\pi/2) - \cos(\tau_{pq}\pi/2)\right| \leq \pi/2|\widehat{\tau}_{pq} - \tau_{pq}|$. Then by definition of $\widehat{\mathbf{M}}^i$ in (4.7) and $\mathbf{M}^i$ in (4.5), for any $p,q = 1,\ldots,d$ we have

$$\left|\widehat{M}^i_{pq} - M^i_{pq}\right| = \left|\pi \cos\left(\frac{\pi}{2}\widehat{\tau}_{pq}\right)\widehat{h}^i_{pq} - \pi \cos\left(\tau_{pq}\frac{\pi}{2}\right)h^i_{pq}\right|$$

$$\leq \pi\left|\cos\left(\frac{\pi}{2}\widehat{\tau}_{pq}\right) - \cos\left(\tau_{pq}\frac{\pi}{2}\right)\right|\left|\widehat{h}^i_{pq}\right| + \pi\left|\cos\left(\frac{\pi}{2}\widehat{\tau}_{pq}\right)\right|\left|\widehat{h}^i_{pq} - h^i_{pq}\right|$$

$$\leq \pi^2|\widehat{\tau}_{pq} - \tau_{pq}| + \pi|\widehat{h}^i_{pq} - h^i_{pq}|,$$

where we use the fact that $|\widehat{h}^i_{pq}| \leq 2$ in the second inequality. Next we have

$$|\widehat{h}^i_{pq} - h^i_{pq}| \leq \Big|\underbrace{\frac{1}{n-1}\sum_{i'\neq i}\operatorname{sign}\big((X_{ip}-X_{i'p})(X_{iq}-X_{i'q})\big)}_{\widehat{m}^i_{pq}} - \underbrace{\mathbb{E}\big[\operatorname{sign}\big((X_{ip}-X_{i'p})(X_{iq}-X_{i'q})\big)\big]}_{m^i_{pq}}\Big|$$

$$+ |\widehat{\tau}_{pq} - \tau_{pq}|,$$

which immediately implies

$$\left|\widehat{M}^i_{pq} - M^i_{pq}\right| \leq (\pi^2 + \pi)|\widehat{\tau}_{pq} - \tau_{pq}| + \pi|\widehat{m}^i_{pq} - m^i_{pq}|. \tag{A.2}$$

Note that $\widehat{m}^i_{pq}$ are the sum of i.i.d. terms conditional on $\boldsymbol{X}_i$, and by Hoeffding's inequality we have

$$\mathbb{P}\big(|\widehat{m}^i_{pq} - m^i_{pq}| > t|\boldsymbol{X}_i\big) \leq 2\exp\Big\{-\frac{(n-1)t^2}{2}\Big\}.$$

It follows that the same inequality holds also unconditionally, and thus by the union bound we obtain

$$\mathbb{P}\big(\max_{i,p,q}|\widehat{m}^i_{pq} - m^i_{pq}| > t\big) \leq 2nd^2\exp\Big\{-\frac{(n-1)t^2}{2}\Big\}.$$

Take $t = 4\sqrt{\log(nd)/n}$, then we have $\max_{i,p,q}|\widehat{m}^i_{pq} - m^i_{pq}| = O_p(\sqrt{\log(nd)/n})$. Recall the Hoeffding's inequality for $\widehat{\tau}_{pq}$ in (6.7), where we have $|\widehat{\tau}_{pq} - \tau_{pq}| = O_p(\sqrt{\log d/n})$. Submitting these two results into (A.2), we obtain

$$\max_{i=1,\ldots,n}\|\widehat{\mathbf{M}}^i - \mathbf{M}^i\|_{\infty,\infty} = O_p(\sqrt{\log(nd)/n}),$$

which immediately implies

$$I_{21} = \|\boldsymbol{\Theta}^*_{*j}\|_1^2 \cdot \|\boldsymbol{\Theta}^*_{*k}\|_1^2 \cdot O_p(\sqrt{\log(nd)/n}) = O_p(M^4\sqrt{\log(nd)/n}) = o_p(1),$$

where the last equality is due to assumption (5.8). Finally, we deal with term $I_{22}$ in (A.1). By Markov's inequality and the finite variance assumption, we have

$$I_{22} = \left|\frac{1}{n}\sum_{i=1}^n \big(\boldsymbol{\Theta}^{*\top}_{*j}\mathbf{M}^i\boldsymbol{\Theta}^*_{*k}\big)^2 - \mathbb{E}\big(\boldsymbol{\Theta}^{*\top}_{*j}\mathbf{M}^i\boldsymbol{\Theta}^*_{*k}\big)^2\right| = O_p\big(\operatorname{Var}\big((\boldsymbol{\Theta}^{*\top}_{*j}\mathbf{M}^i\boldsymbol{\Theta}^*_{*k})^2/n\big)\big).$$

Therefore, in order to prove $I_{22} = o(1)$, it suffices to show that $\operatorname{Var}\big((\boldsymbol{\Theta}^{*\top}_{*j}\mathbf{M}^i\boldsymbol{\Theta}^*_{*k})^2 = o_p(n)$. In fact, we have $|\boldsymbol{\Theta}^{*\top}_{*j}\mathbf{M}^i\boldsymbol{\Theta}^*_{*k}| \leq M^2 \cdot 2\pi$. And by Assumption 5.1 we have $\|\boldsymbol{\Theta}^*_{*j}\|_1 \leq M$, and by the assumption (5.7) we have that $M^4 \cdot O_p(s\sqrt{\log d/n}) = o_p(1)$, then it follows that $\|\boldsymbol{\Theta}^*_{*j}\|_1^4 \cdot \|\boldsymbol{\Theta}^*_{*k}\|_1^4 = o_p(n)$. Thus we have $I_{22} = o_p(1)$, and this completes the proof. $\square$

Now we are going to prove Lemma 6.1, which is useful in the proof of main theory in Section 6. We first lay out the following lemma that shows the average estimation error on each machine.

**Lemma A.1.** Given $X_1, X_2, \ldots, X_N$ are i.i.d. random vectors following $TE_d(\Sigma^*, \xi; f_1, f_2, \ldots, f_d)$. Let $\widehat{\Sigma}^{(l)}$ be the Kendall's tau correlation matrix of $X_{(l-1)n+1}, \cdots, X_{ln}$, where $l = 1, \cdots, m$ and $N = nm$. Then we have

$$\left\| \frac{1}{m} \sum_{l=1}^{m} (\widehat{\Sigma}^{(l)} - \Sigma^*) \right\|_{\infty,\infty} \leq 4\sqrt{\frac{\log d}{N}} + \sqrt{\frac{d}{N(n-1)}}$$

holds with probability at least $1 - 74/d$.

*Proof of Lemma 6.1.* By definition in (6.1), we have

$$\widetilde{\Theta}^{(l)} - \Theta^* = 2\widehat{\Theta}^{(l)} - \widehat{\Theta}^{(l)} \widehat{\Sigma}^{(l)} \widehat{\Theta}^{(l)} - \Theta^* = -\Theta^* W^{(l)} \Theta^* + \text{Rem}^{(l)}, \tag{A.3}$$

where $\text{Rem}^{(l)} = -(\widehat{\Theta}^{(l)} - \Theta^*) W^{(l)} \Theta^* - (\widehat{\Theta}^{(l)} \widehat{\Sigma}^{(l)} - I)(\widehat{\Theta}^{(l)} - \Theta^*)$ for $l = 1, \ldots, m$. Therefore, we have

$$\|\bar{\Theta} - \Theta^*\|_{\infty,\infty} \leq \left\| \frac{1}{m} \sum_{l=1}^{m} \Theta^* W^{(l)} \Theta^* \right\|_{\infty,\infty} + \frac{1}{m} \sum_{l=1}^{m} \|\text{Rem}^{(l)}\|_{\infty,\infty}.$$

In (6.4), we have proved that

$$\|\text{Rem}^{(l)}\|_{\infty,\infty} \leq 6\pi C_1 M^3 \frac{s \log d}{n} + 4C_0 C_1 M^4 K_{\Sigma^*} \frac{s \log d}{n} + 3\pi C_1^2 M^4 s^2 \left(\frac{\log d}{n}\right)^{3/2} \tag{A.4}$$

holds with probability at least $1 - d^{-1} - d^{-5/2}$. Note that the above inequality holds for any $l = 1, \ldots, m$. And by Lemma A.1, with probability at least $1 - 74/d$ we have

$$\left\| \frac{1}{m} \sum_{l=1}^{m} \Theta^* W^{(l)} \Theta^* \right\|_{\infty,\infty} \leq M^2 \left\| \frac{1}{m} \sum_{l=1}^{m} (\widehat{\Sigma}^{(l)} - \Sigma^*) \right\|_{\infty,\infty} \leq 4M^2 \sqrt{\frac{\log d}{N}} + M^2 \sqrt{\frac{d}{N(n-1)}}.$$

Therefore we obtain

$$\|\bar{\Theta} - \Theta^*\|_{\infty,\infty} \leq 4M^2 \sqrt{\frac{\log d}{N}} + M^2 \sqrt{\frac{d}{N(n-1)}}$$
$$+ 6\pi C_1 M^3 \frac{s \log d}{n} + 4C_0 C_1 M^4 K_{\Sigma^*} \frac{s \log d}{n} + 3\pi C_1^2 M^4 s^2 \left(\frac{\log d}{n}\right)^{3/2}$$
$$\leq 4M^2 \sqrt{\frac{\log d}{N}} + M^2 \sqrt{\frac{d}{N(n-1)}}$$
$$+ \left( 6\pi C_1 M^3 + 4C_0 C_1 M^4 K_{\Sigma^*} + 3\pi C_1^2 M^4 s \sqrt{\frac{\log d}{n}} \right) \frac{s \log d}{n}$$
$$\leq 4M^2 \sqrt{\frac{\log d}{N}} + M^2 \sqrt{\frac{d}{N(n-1)}} + CM^4 K_{\Sigma^*} \frac{sm \log d}{N}$$

holds with probability at least $1 - 76/d$, where $C$ is an absolute constant and the last inequality is due to the fact that $s\sqrt{\log d/n} = o_p(1)$ by Lemma D.1. □

# B  Proofs for Gaussian Graphical Models

We first lay out some lemmas which are useful in proving Theorem 7.1.

**Lemma B.1.** Given $\boldsymbol{X}_1, \boldsymbol{X}_2, \ldots, \boldsymbol{X}_N$ being i.i.d. sub-Gaussian random variables with $\text{Var}(\boldsymbol{X}_i) = \boldsymbol{\Sigma}^*$, and suppose $\|\boldsymbol{X}_i\|_{\psi_2} \leq \kappa$. Let $\widehat{\boldsymbol{\Sigma}}^{(l)} = 1/n \sum_{i_l=(l-1)n+1}^{ln} \boldsymbol{X}_{i_l} \boldsymbol{X}_{i_l}^\top$ be the covariance matrix of $\boldsymbol{X}_{(l-1)n+1}, \cdots, \boldsymbol{X}_{ln}$, where $l = 1, \cdots, m$ and $N = nm$. Then we have

$$\left\| \frac{1}{m} \sum_{l=1}^{m} (\widehat{\boldsymbol{\Sigma}}^{(l)} - \boldsymbol{\Sigma}^*) \right\|_{\infty,\infty} \leq \kappa^2 \sqrt{\frac{\log d}{N}},$$

holds with probability at least $1 - 1/d^{C_2-2}n$ where $C_2 > 0$ is an absolute constant.

The following lemma is the equivalent result for Gaussian graphical models to Lemma D.1.

**Lemma B.2.** For the $n$ data on any machine, say $\boldsymbol{X}_1, \cdots, \boldsymbol{X}_n$, suppose that the covariance matrix $\boldsymbol{\Sigma}^*$ satisfies Assumptions 5.1 and 5.2, the precision matrix $\boldsymbol{\Theta}^* \in \mathcal{U}(s, M)$ defined in Section 5, and $\lambda \geq M \|\widehat{\boldsymbol{\Sigma}} - \boldsymbol{\Sigma}^*\|_{\infty,\infty} = C\nu\sqrt{\log d/n}$ for some constant $C_0 > 0$. Then with probability at least $1 - d^{-1}$, we have

$$\|\widehat{\boldsymbol{\Theta}} - \boldsymbol{\Theta}^*\|_{\infty,\infty} \leq 4C_0 M^2 \sqrt{\frac{\log d}{n}}, \qquad \|\widehat{\boldsymbol{\Theta}} - \boldsymbol{\Theta}^*\|_1 \leq C_1 M^2 s \sqrt{\frac{\log d}{n}},$$

where $C_1$ is a constant only depending on $C_0$.

Based on the two lemmas above, we also have the following Lemma B.3.

**Lemma B.3.** The average of debiased estimators on all the machines $\bar{\boldsymbol{\Theta}}$ satisfies

$$\|\bar{\boldsymbol{\Theta}} - \boldsymbol{\Theta}^*\|_{\infty,\infty} \leq M^2 K_{\boldsymbol{\Sigma}^*} \sqrt{\frac{\log d}{N}} + C' M^4 K_{\boldsymbol{\Sigma}^*} \frac{s \log d}{n}.$$

which holds with probability at least $1 - 3/d$, where $C'$ is an absolute constant.

Now we are ready to prove the main theories in Section 7.

*Proof of Theorem 7.1.* By assumption and Lemma B.3, we have $\|\bar{\boldsymbol{\Theta}} - \boldsymbol{\Theta}^*\|_{\infty,\infty} \leq t$ with probability at least $1 - 2/d - 1/d^{C_2-2}$, where $C_2 > 2$ is an absolute constant. Recall the definition of hard thresholding function, we have that $\text{supp}(\check{\boldsymbol{\Theta}})$ is the set of index $(j, k)$ such that $\bar{\Theta}_{jk} > t$. Thus, for $(j, k) \in \text{supp}(\check{\boldsymbol{\Theta}})$, we have $\check{\Theta}_{jk} = \bar{\Theta}_{jk}$; for $(j, k) \notin \text{supp}(\check{\boldsymbol{\Theta}})$, we have $|\check{\Theta}_{jk} - \bar{\Theta}_{jk}| = |0 - \bar{\Theta}_{jk}| \leq t$. Therefore, we obtain $\|\check{\boldsymbol{\Theta}} - \bar{\boldsymbol{\Theta}}\|_{\infty,\infty} \leq t$. By triangle inequality,

$$\|\check{\boldsymbol{\Theta}} - \boldsymbol{\Theta}^*\|_{\infty,\infty} \leq \|\check{\boldsymbol{\Theta}} - \bar{\boldsymbol{\Theta}}\|_{\infty,\infty} + \|\bar{\boldsymbol{\Theta}} - \boldsymbol{\Theta}^*\|_{\infty,\infty}$$

$$\leq t + t = 2M^2 K_{\boldsymbol{\Sigma}^*} \sqrt{\frac{\log d}{N}} + 2C' M^4 K_{\boldsymbol{\Sigma}^*} \frac{s \log d}{n}$$

holds with probability at least $1 - 2/d - 1/d^{C_2-2}$ by Lemma B.3. Since by Assumption 5.1, $\boldsymbol{\Theta}^* \in \mathcal{U}(s, M)$, we have the row sparsity claim: $\|\boldsymbol{\Theta}^*\|_{\infty,0} \leq s$. When $t > \|\bar{\boldsymbol{\Theta}} - \boldsymbol{\Theta}^*\|_{\infty,\infty}$, we have $\check{\Theta}_{jk} = 0$ whenever $\Theta^*_{jk} = 0$ due to the thresholding function, which implies that $\|\check{\boldsymbol{\Theta}} - \boldsymbol{\Theta}^*\|_{\infty,0} \leq s$.

By properties of matrix norms, we have $\|\boldsymbol{\Theta}\|_2^2 \leq \|\boldsymbol{\Theta}\|_1 \cdot \|\boldsymbol{\Theta}\|_\infty$ and $\|\boldsymbol{\Theta}\|_1 = \|\boldsymbol{\Theta}\|_\infty$ when $\boldsymbol{\Theta}$ is symmetric. Then with probability at least $1 - 2/d - 1/d^{C_2-2}$, we have

$$\|\check{\boldsymbol{\Theta}} - \boldsymbol{\Theta}^*\|_2 \leq \|\check{\boldsymbol{\Theta}} - \boldsymbol{\Theta}^*\|_1 \leq s \cdot \|\check{\boldsymbol{\Theta}} - \boldsymbol{\Theta}^*\|_{\infty,\infty} \leq 2M^2 K_{\boldsymbol{\Sigma}^*} s \sqrt{\frac{\log d}{N}} + 2C'M^4 K_{\boldsymbol{\Sigma}^*} \frac{s^2 \log d}{n},$$

$$\|\check{\boldsymbol{\Theta}} - \boldsymbol{\Theta}^*\|_F \leq \sqrt{sd}\|\check{\boldsymbol{\Theta}} - \boldsymbol{\Theta}^*\|_{\infty,\infty} \leq 2M^2 K_{\boldsymbol{\Sigma}^*} \sqrt{sd\frac{\log d}{N}} + 2C'M^4 K_{\boldsymbol{\Sigma}^*} \frac{s^{3/2}\sqrt{d}\log d}{n},$$

where $C', C_2 > 0$ are absolute constants. $\square$

Next we prove the result for inference of Gaussian graphical models.

*Proof of Theorem 7.3.* For the average of debiased estimators $\bar{\boldsymbol{\Theta}}$, by definition in (6.1) we have

$$\bar{\boldsymbol{\Theta}} - \boldsymbol{\Theta}^* = \frac{1}{m}\sum_{l=1}^{m} \widetilde{\boldsymbol{\Theta}}^{(l)} - \boldsymbol{\Theta}^* = -\frac{1}{m}\sum_{l=1}^{m}\left(\boldsymbol{\Theta}^*\widehat{\boldsymbol{\Sigma}}^{(l)}\boldsymbol{\Theta}^* - \boldsymbol{\Theta}^* + \text{Rem}^{(l)}\right).$$

In (6.4), we have proved that $\|\text{Rem}^{(l)}\|_{\infty,\infty} = O_p(s\log d/n)$. Recall that $\mathbf{W}^{(l)} := \widehat{\boldsymbol{\Sigma}}^{(l)} - \boldsymbol{\Sigma}^*$, and $\widehat{\boldsymbol{\Sigma}}^{(l)} = 1/n \sum_{i=1}^n \boldsymbol{X}_{(l-1)n+i}\boldsymbol{X}_{(l-1)n+i}^\top$ for $l = 1, \cdots, m$. Hence for every $j, k = 1, \cdots, d$ it holds that

$$\sqrt{n}(\bar{\Theta}_{jk} - \Theta^*_{jk}) = \frac{1}{m}\sum_{l=1}^m \sqrt{n}\left(\widetilde{\Theta}^{(l)}_{jk} - \Theta^*_{jk}\right)$$

$$= \frac{1}{m}\sum_{l=1}^m \sqrt{n}\left(-\boldsymbol{\Theta}^*_{j*}\mathbf{W}^{(l)}\boldsymbol{\Theta}^*_{*k} + \left[\text{Rem}^{(l)}\right]_{jk}\right)$$

$$= -\frac{1}{m\sqrt{n}}\sum_{i=1}^n\sum_{l=1}^m \left(\boldsymbol{\Theta}^*_{j*}\boldsymbol{X}_{(l-1)n+i}\boldsymbol{X}^\top_{(l-1)n+i}\boldsymbol{\Theta}^*_{*k} - \Theta^*_{jk}\right) + o_p(1),$$

where we used the fact that $\sqrt{n}\left[\text{Rem}^{(l)}\right]_{jk} = o_p(1)$ when $n > (s\log d)^2$. We need to show the first term above weakly converges to the normal distribution. To this end, define

$$Z_{jk,i} = \frac{1}{m}\sum_{l=1}^m \left(\boldsymbol{\Theta}^*_{j*}\boldsymbol{X}_{(l-1)n+i}\boldsymbol{X}^\top_{(l-1)n+i}\boldsymbol{\Theta}^*_{*k} - \Theta^*_{jk}\right) = \frac{1}{m}\sum_{l=1}^m \left(\boldsymbol{\Theta}^*_{j*}\boldsymbol{X}_{(l-1)n+i}\boldsymbol{\Theta}^*_{k*}\boldsymbol{X}_{(l-1)n+i} - \Theta^*_{jk}\right),$$

for each $i = 1, \cdots, n, l = 1, \cdots, m$, since each $\boldsymbol{X}_{(l-1)n+i}$ has sub-Gaussian elements, the variance $\sigma^2_{jk} = \text{var}(Z_{jk,i}) = 1/m \cdot \text{var}\left(\boldsymbol{\Theta}^*_{j*}\boldsymbol{X}_{(l-1)n+i}\boldsymbol{\Theta}^*_{k*}\boldsymbol{X}_{(l-1)n+i}\right)$ is finite. Actually, since $\boldsymbol{X}_1 \sim N(\boldsymbol{0}, \boldsymbol{\Sigma}^*)$, then $\boldsymbol{\Theta}^*\boldsymbol{X}_1 \sim N(\boldsymbol{0}, \boldsymbol{\Theta}^*)$, and

$$\text{var}(\boldsymbol{\Theta}^*_{j*}\boldsymbol{X}_1\boldsymbol{\Theta}^*_{k*}\boldsymbol{X}_1) = \text{var}(\boldsymbol{\Theta}^*_{j*}\boldsymbol{X}_1\boldsymbol{X}_1^\top\boldsymbol{\Theta}^{*\top}_{k*}) = \text{var}(e_i^\top ZZ^\top e_j),$$

where $e_i, e_j$ are natural basis in $\mathbb{R}^d$ and $Z \sim N(\boldsymbol{0}, \boldsymbol{\Theta}^*)$. Thus

$$m\sigma^2_{jk} = \text{var}(Z^j Z^k) = \mathbb{E}((Z^j Z^k)^2) - (\mathbb{E}(Z^j Z^k))^2 = \Theta^*_{jj}\Theta^*_{kk} + 2{\Theta^*_{jk}}^2 - {\Theta^*_{jk}}^2 = \Theta^*_{jj}\Theta^*_{kk} + {\Theta^*_{jk}}^2.$$

By Assumptions 5.1 and 5.2, we know $\sigma^2_{jk}$ is finite.

Now we have known that $Z_{jk,i}$ are identically independently distributed r.v.s with $\mathbb{E}(Z_{jk,i}) = 1/m \sum_{l=1}^{m} \left( \boldsymbol{\Theta}_{j*}^* \boldsymbol{\Sigma}^* \boldsymbol{\Theta}_{*k}^* - \Theta_{jk}^* \right) = 0$ and $\text{var}(Z_{jk,i}) = \sigma_{jk}^2$. Denote $S_n = \sum_{i=1}^{n} Z_{jk,i}$, and $s_n^2 := \text{var}(S_n) = n\sigma_{jk}^2$. We obtain

$$\frac{\sqrt{n}}{m} \sum_{l=1}^{m} \frac{\boldsymbol{\Theta}_{j*}^* \widehat{\boldsymbol{\Sigma}}^{(l)} \boldsymbol{\Theta}_{*k}^* - \Theta_{jk}^*}{\sigma_{jk}} = \frac{S_n}{s_n} + \frac{o_p(1)}{\sigma_{jk}} = Z_{jk}^n + \frac{o_p(1)}{\sigma_{jk}}, \tag{B.1}$$

where $\widehat{\boldsymbol{\Sigma}}^{(l)} = 1/n \sum_{i=1}^{n} \boldsymbol{X}_{(l-1)n+i} \boldsymbol{X}_{(l-1)n+i}^\top$. Note that $1/\sigma_{jk} = O(1)$, we have $o_p(1)/\sigma_{jk} = o_p(1)$. In order to show $S_n/s_n \rightsquigarrow N(0,1)$, we now verify the Lyapunovs condition for CLT:

$$\lim_{n \to \infty} \frac{1}{s_n^{2+\delta}} \sum_{i=1}^{n} \mathbb{E}|Z_{jk,i}|^{2+\delta} = 0.$$

Here we set $\delta = 1$. To simplify the notation, we denote $Z_{jk,i} = 1/m \sum_{l=1}^{m} Z_{jk,i}^l$. Clearly, we have that $Z_{jk,i}^l$ are i.i.d. for all $l = 1, \ldots, m$ and $\mathbb{E}Z_{jk,i}^l = 0$. Therefore, it implies that $\mathbb{E}|Z_{jk,i}|^3 = 1/m^3 \sum_{l=1}^{m} \mathbb{E}|Z_{jk,i}^l|^3$. Since $\|\boldsymbol{\Theta}_{j*}^*\|_2 \leq \|\boldsymbol{\Theta}^*\|_2 \leq \nu$ and $\|\boldsymbol{X}_1\|_{\psi_2} \leq \|\boldsymbol{\Sigma}^*\|_2 \leq \nu$, by Lemma D.2, we have

$$\mathbb{E}|Z_{jk,i}|^3 = \frac{1}{m^3} \sum_{l=1}^{m} \mathbb{E}|Z_{jk,i}^l|^3 \leq \frac{24\nu^{12}}{m^2}.$$

It follows that

$$\frac{1}{s_n^3} \sum_{i=1}^{n} \mathbb{E}|Z_{jk,i}|^3 \leq \frac{m^{3/2}}{n^{3/2}(\Theta_{jj}^* \Theta_{kk}^* + \Theta_{jk}^{*2})^{3/2}} \cdot \frac{24n\nu^{12}}{m^2} = \frac{24\nu^{12}}{N^{1/2}(\Theta_{jj}^* \Theta_{kk}^* + \Theta_{jk}^{*2})^{3/2}} = o(1).$$

Thus we have proved $S_n/s_n \xrightarrow{d} N(0,1)$ and then $\sqrt{n}/\sqrt{m} \sum_{l=1}^{m} (\boldsymbol{\Theta}_{j*}^* \widehat{\boldsymbol{\Sigma}}^{(l)} \boldsymbol{\Theta}_{*k}^* - \Theta_{jk}^*)/\sqrt{\Theta_{jj}^* \Theta_{kk}^* + \Theta_{jk}^{*2}} \xrightarrow{d} N(0,1)$ by (B.1). Recalling the plugging process in Section 4.3, by Theorem B.3 we already have $\widehat{\boldsymbol{\Theta}}^{(l)} \xrightarrow{p} \boldsymbol{\Theta}^*$, and then by Slutsky's theorem we obtain

$$\frac{\sqrt{N}}{m} \sum_{l=1}^{m} \frac{\widehat{\boldsymbol{\Theta}}_{j*} \widehat{\boldsymbol{\Sigma}}^{(l)} \widehat{\boldsymbol{\Theta}}_{*k} - \Theta_{jk}^*}{\sqrt{\widehat{\Theta}_{jj}^{(l)} \widehat{\Theta}_{kk}^{(l)} + (\widehat{\Theta}_{jk}^{(l)})^2}} \rightsquigarrow N(0,1).$$

□

## C  Proofs of Additional Lemmas

In this section, we are going to prove the additional lemmas used in the proof of both transelliptical graphical models and Gaussian graphical models.

### C.1  Proof of Additional Lemma for Transelliptical Graphical Models

We first prove Lemma A.1, which is used in the proof of Lemma 6.1.

*Proof of Lemma A.1.* By definition, we have

$$\left\|\frac{1}{m}\sum_{l=1}^{m}(\widehat{\boldsymbol{\Sigma}}^{(l)} - \boldsymbol{\Sigma}^*)\right\|_{\infty,\infty} = \sup_{p,q}\left|\frac{1}{m}\sum_{l=1}^{m}(\widehat{\boldsymbol{\Sigma}}_{pq}^{(l)} - \boldsymbol{\Sigma}_{pq}^*)\right|.$$

For any $p, q = 1, \cdots, d$, by Taylor's expansion of $\sin(x)$, we have

$$\frac{1}{m}\sum_{l=1}^{m}(\widehat{\boldsymbol{\Sigma}}_{pq}^{(l)} - \boldsymbol{\Sigma}_{pq}^*) = \frac{1}{m}\sum_{l=1}^{m}\left(\sin\left(\frac{\pi}{2}\widehat{\tau}_{pq}^{(l)}\right) - \sin\left(\frac{\pi}{2}\tau_{pq}\right)\right)$$

$$= \frac{1}{m}\sum_{l=1}^{m}\left[\cos\left(\frac{\pi}{2}\tau_{pq}\right)\frac{\pi}{2}(\widehat{\tau}_{pq}^{(l)} - \tau_{pq}) - \frac{1}{2}\sin\left(\frac{\pi}{2}\widetilde{\tau}_{pq}^{(l)}\right)\left(\frac{\pi}{2}(\widehat{\tau}_{pq}^{(l)} - \tau_{pq})\right)^2\right], \quad \text{(C.1)}$$

where $\widetilde{\tau}_{pq}^{(l)}$ is a number between $\widehat{\tau}_{pq}^{(l)}$ and $\tau_{pq}$. We first consider $\widehat{\tau}_{pq}^{(l)} - \tau_{pq}$, $p, q = 1, \cdots, d$. For simpleness, we leave out the index $l$ for the time being. Recall the notations for Hájek's projection defined in (4.4). We have

$$\widehat{\tau}_{pq} - \tau_{pq} = \frac{2}{n}\sum_{i=1}^{n} h_{pq}^{i} + \frac{2}{n(n-1)}\sum_{1\leq i<i'\leq n} w_{pq}^{ii'}.$$

And thus for the second term on the right hand of (C.1), we have

$$\frac{1}{m}\sum_{l=1}^{m}\left[\cos\left(\frac{\pi}{2}\tau_{pq}\right)(\widehat{\tau}_{pq}^{(l)} - \tau_{pq})\right]$$

$$= \frac{2}{mn}\sum_{l=1}^{m}\sum_{i_l=1}^{n}\cos\left(\frac{\pi}{2}\tau_{pq}\right)h_{pq}^{i_l} + \frac{2}{mn(n-1)}\sum_{l=1}^{m}\sum_{1\leq i_l<i_l'\leq n}\cos\left(\frac{\pi}{2}\tau_{pq}\right)w_{pq}^{i_l i_l'}$$

$$= \underbrace{\frac{2}{N}\left(\sum_{i_1=1}^{n}\cos\left(\frac{\pi}{2}\tau_{pq}\right)h_{pq}^{i_1} + \cdots + \sum_{i_m=(m-1)n+1}^{N}\cos\left(\frac{\pi}{2}\tau_{pq}\right)h_{pq}^{i_m}\right)}_{I_1}$$

$$+ \underbrace{\frac{2}{N(n-1)}\left(\sum_{1\leq i_1<i_1'\leq n}\cos\left(\frac{\pi}{2}\tau_{pq}\right)w_{pq}^{i_1 i_1'} + \cdots + \sum_{(m-1)n+1\leq i_m<i_m'\leq N}\cos\left(\frac{\pi}{2}\tau_{pq}\right)w_{pq}^{i_m i_m'}\right)}_{I_2}.$$

By writing it as two terms, we want to bound $I_1$ and show that $I_2$ is asymptotically a small term. By definition of $h_{pq}^{i_l}$ and independence of $m$ machines, we know they are independent for every $i = 1, \cdots, N$. Note that $|h_{pq}^{i_l}| \leq 2$ and $\mathbb{E}h_{pq}^{i_l} = 0$, in McDiarmid's inequality (D.5), define $f$ as follows:

$$f(h_{pq}^1, \ldots, h_{pq}^N) = \frac{2}{N}\left(\sum_{i_1=1}^{n}\cos\left(\frac{\pi}{2}\tau_{pq}\right)h_{pq}^{i_1} + \cdots + \sum_{i_m=(m-1)n+1}^{N}\cos\left(\frac{\pi}{2}\tau_{pq}\right)h_{pq}^{i_m}\right).$$

It is easy to check that for any $i = 1, \ldots, N$, if $h_{pq}^i, h_{pq}^{i'}$ are different sample variables drawn from $h_{pq}^i$, then we have

$$\sup_{h_{pq}^1,\ldots,h_{pq}^N,h_{pq}^{i'}}\left|f(h_{pq}^1,\ldots,h_{pq}^i,\ldots,h_{pq}^N) - f(h_{pq}^1,\ldots,h_{pq}^{i'},\ldots,h_{pq}^N)\right| \leq \frac{4}{N}.$$

It follows by applying McDiarmid's inequality in Lemma D.5 to obtain

$$\mathbb{P}\Big(\big|f(h_{pq}^1,\ldots,h_{pq}^N) - \mathbb{E}(f(h_{pq}^1,\ldots,h_{pq}^N))\big| \geq t\Big) \leq 2\exp\Big(-\frac{Nt^2}{8}\Big),$$

take $t = 2\sqrt{2\log d/N}$, and note that $\mathbb{E}(f(h_{pq}^1,\ldots,h_{pq}^N)) = 0$ we get

$$I_1 \leq 4\sqrt{\frac{2\log d}{N}},$$

with probability at least $1 - 2/d$.

Next we deal with $I_2$. We already know that $w_{pq}^{i_l i_l'}$ are independent for every $l = 1,\cdots,m$. By the claim in (6.6) and note that $\mathbb{E} w_{pq}^{i_l i_l'} = 0$, we have

$$\text{Var}(I_2) = \frac{4}{N^2(n-1)^2}\sum_{l=1}^m \cos\Big(\frac{\pi}{2}\tau_{pq}\Big)^2 \text{Var}\Big(\sum_{(l-1)n+1\leq i_l < i_l' \leq ln} w_{pq}^{i_l i_l'}\Big)$$

$$= \frac{4}{N^2(n-1)^2}\sum_{l=1}^m \cos\Big(\frac{\pi}{2}\tau_{pq}\Big)^2 \sum_{(l-1)n+1\leq i_l < i_l' \leq ln} \mathbb{E}\big(w_{pq}^{i_l i_l'}\big)^2$$

$$\leq \frac{4}{N^2(n-1)^2} \cdot m \cdot \frac{n(n-1)}{2} \cdot 6^2 = \frac{72}{N(n-1)},$$

where we used the fact that $|w_{pq}^{ii'}| \leq |h_{pq}^{ii'}| + |h_{pq}^{ii'|i}| + |h_{pq}^{ii'|i'}| \leq 6$. By Chebyshev's inequality

$$\mathbb{P}(|I_2| \geq t) \leq \frac{\text{Var}(I_2)}{t^2}.$$

Let $t = \sqrt{d/(N(n-1))}$, then we obtain

$$|I_2| \leq \sqrt{\frac{d}{N(n-1)}},$$

holds with probability at least $1 - 72/d$. Then we obtain

$$\Big|\frac{1}{m}\sum_{l=1}^m \cos\Big(\frac{\pi}{2}\tau_{pq}\Big)(\widehat{\tau}_{pq}^{(l)} - \tau_{pq})\Big| \leq |I_1| + |I_2| \leq 4\sqrt{\frac{\log d}{N}} + \sqrt{\frac{d}{N(n-1)}}$$

holds with probability at least $1 - 74/d$. Now let us come back to (C.1), for any $l = 1,\ldots,m$, we have taylor expansion

$$\sin\Big(\frac{\pi}{2}\widehat{\tau}_{pq}^{(l)}\Big) - \sin\Big(\frac{\pi}{2}\tau_{pq}\Big) = \cos\Big(\frac{\pi}{2}\tau_{pq}\Big)\frac{\pi}{2}(\widehat{\tau}_{pq}^{(l)} - \tau_{pq}) + R_2,$$

where $R_2$ is the remainder

$$R_2 = -\frac{1}{2}\sin\Big(\frac{\pi}{2}\widetilde{\tau}_{pq}^{(l)}\Big)\Big(\frac{\pi}{2}(\widehat{\tau}_{pq}^{(l)} - \tau_{pq})\Big)^2.$$

When $N$ goes to infinity, we have $\widehat{\tau}_{pq}^{(l)} \to \tau_{pq}$, and $R_2 = o(\widehat{\tau}_{pq}^{(l)} - \tau_{pq})$, namely $R_2$ is small term of higher order than $\cos(\pi/2\tau_{pq})(\widehat{\tau}_{pq}^{(l)} - \tau_{pq})$. Thus the second term in (C.1) is controlled by the first term, and it follows that

$$\left\|\frac{1}{m}\sum_{l=1}^{m}(\widehat{\boldsymbol{\Sigma}}^{(l)} - \boldsymbol{\Sigma}^*)\right\|_{\infty,\infty} = \sup_{p,q}\left|\frac{1}{m}\sum_{l=1}^{m}(\widehat{\boldsymbol{\Sigma}}_{pq}^{(l)} - \boldsymbol{\Sigma}_{pq}^*)\right| \leq 4\sqrt{\frac{\log d}{N}} + \sqrt{\frac{d}{N(n-1)}}$$

holds with probability at least $1 - 74/d$. $\square$

## C.2 Proofs of Additional Lemmas for Gaussian Graphical Models

Now we prove the lemmas used in the proof of Gaussian graphical models.

*Proof of Lemma B.1.* First we show that

$$\frac{1}{m}\sum_{l=1}^{m}(\widehat{\boldsymbol{\Sigma}}^{(l)} - \boldsymbol{\Sigma}^*) = \frac{1}{m}\sum_{l=1}^{m}\left(\frac{1}{n}\sum_{i_l=(l-1)n+1}^{ln} \boldsymbol{X}_{i_l}\boldsymbol{X}_{i_l}^\top - \boldsymbol{\Sigma}^*\right) = \frac{1}{N}\sum_{i=1}^{N}(\boldsymbol{X}_i\boldsymbol{X}_i^\top - \boldsymbol{\Sigma}^*).$$

Let $X^{(p)}$ denote the $p$-th element of $\boldsymbol{X}$, $p = 1, \cdots, d$. Since $\boldsymbol{X}_i$ is a sub-Gaussian random vector, then each element of it $X_i^{(p)}$ is a sub-Gassian variable, $i = 1, \cdots, N$. Note that $\mathbb{E}(\boldsymbol{X}_1\boldsymbol{X}_1^\top) = \boldsymbol{\Sigma}_0$ and by Lemma D.3 we know $X_i^{(p)}X_i^{(q)} - \Sigma_{pq}^*$ is a centered sub-exponential random variable. By Lemma D.4 we have

$$\mathbb{P}\left(\left\|\frac{1}{m}\sum_{l=1}^{m}(\widehat{\boldsymbol{\Sigma}}^{(l)} - \boldsymbol{\Sigma}^*)\right\|_\infty > t\right) \leq \sum_{p,q=1}^{d} \mathbb{P}\left(\frac{1}{m}\sum_{l=1}^{m}\left(\widehat{\Sigma}_{pq}^{(l)} - \Sigma_{pq}^*\right) > t\right)$$

$$= \sum_{p,q=1}^{d} \mathbb{P}\left(\frac{1}{N}\sum_{i=1}^{N}\left(X_i^{(p)}X_i^{(q)\top} - \Sigma_{pq}^*\right) > t\right)$$

$$\leq d^2\exp\left(-C_2\min\left\{\frac{t^2 N}{\kappa^4}, \frac{tN}{\kappa^2}\right\}\right),$$

where $C_2$ is a constant greater than 2. Taking $t = \kappa^2\sqrt{\log d/N}$, we have

$$\left\|\frac{1}{m}\sum_{l=1}^{m}(\widehat{\boldsymbol{\Sigma}}^{(l)} - \boldsymbol{\Sigma}^*)\right\|_\infty \leq \kappa^2\sqrt{\frac{\log d}{N}},$$

with probability at least $1 - 1/d^{C_2-2}$. $\square$

Next, we prove Lemma B.3.

*Proof of Lemma B.3.* Recall the definition in (6.1), we have $\mathbf{W}^{(l)} := \widehat{\boldsymbol{\Sigma}}^{(l)} - \boldsymbol{\Sigma}^*$ and

$$\widetilde{\boldsymbol{\Theta}}^{(l)} - \boldsymbol{\Theta}^* = 2\widehat{\boldsymbol{\Theta}}^{(l)} - \widehat{\boldsymbol{\Theta}}^{(l)}\widehat{\boldsymbol{\Sigma}}^{(l)}\widehat{\boldsymbol{\Theta}}^{(l)} - \boldsymbol{\Theta}^* = -\boldsymbol{\Theta}^*\mathbf{W}^{(l)}\boldsymbol{\Theta}^* + \text{Rem}^{(l)},$$

where $\text{Rem}^{(l)} = -(\widehat{\boldsymbol{\Theta}}^{(l)} - \boldsymbol{\Theta}^*)\mathbf{W}^{(l)}\boldsymbol{\Theta}^* - (\widehat{\boldsymbol{\Theta}}^{(l)}\widehat{\boldsymbol{\Sigma}}^{(l)} - \mathbf{I})(\widehat{\boldsymbol{\Theta}}^{(l)} - \boldsymbol{\Theta}^*)$. Then by definition and triangle inequality we have

$$\|\bar{\boldsymbol{\Theta}} - \boldsymbol{\Theta}^*\|_{\infty,\infty} = \left\|\frac{1}{m}\sum_{l=1}^m \widetilde{\boldsymbol{\Theta}}^{(l)} - \boldsymbol{\Theta}^*\right\|_{\infty,\infty} = \left\|\frac{1}{m}\sum_{l=1}^m \left(-\boldsymbol{\Theta}^*\mathbf{W}^{(l)}\boldsymbol{\Theta}^* + \text{Rem}^{(l)}\right)\right\|_{\infty,\infty}$$
$$\leq \left\|\frac{1}{m}\sum_{l=1}^m \boldsymbol{\Theta}^*\mathbf{W}^{(l)}\boldsymbol{\Theta}^*\right\|_{\infty,\infty} + \frac{1}{m}\sum_{l=1}^m \|\text{Rem}^{(l)}\|_{\infty,\infty}.$$

From (6.4) we have

$$\|\text{Rem}^{(l)}\|_{\infty,\infty} \leq 6\pi C_1 M^3 s \frac{\log d}{n} + 4C_0 C_1 M^4 s K_{\boldsymbol{\Sigma}^*} \frac{\log d}{n} + 3\pi C_1^2 M^4 s^2 \left(\frac{\log d}{n}\right)^{3/2}$$

holds with probability at least $1 - d^{-1} - d^{-5/2} > 1 - 2/d$. Note that $\boldsymbol{X}_1,\ldots,\boldsymbol{X}_N$ follow multivariate noraml distribution $\mathcal{N}(\boldsymbol{0}, \boldsymbol{\Sigma}^*)$, thus $\|\boldsymbol{X}_i\|_{\psi_2} \leq \sqrt{K_{\boldsymbol{\Sigma}^*}}$ by Assumption 5.2. Applying Lemma B.1 we have

$$\left\|\frac{1}{m}\sum_{l=1}^m \boldsymbol{\Theta}^*\mathbf{W}^{(l)}\boldsymbol{\Theta}^*\right\|_{\infty,\infty} \leq M^2 \left\|\frac{1}{m}\sum_{l=1}^m (\widehat{\boldsymbol{\Sigma}}^{(l)} - \boldsymbol{\Sigma}^*)\right\|_{\infty,\infty} \leq M^2 K_{\boldsymbol{\Sigma}^*}\sqrt{\frac{\log d}{N}}$$

holds with probability at least $1 - 1/d^{C_2-2}$. Thus we have

$$\|\bar{\boldsymbol{\Theta}} - \boldsymbol{\Theta}^*\|_{\infty,\infty} \leq M^2 K_{\boldsymbol{\Sigma}^*}\sqrt{\frac{\log d}{N}} + 6\pi C_1 M^3 s \frac{\log d}{n} + 4C_0 C_1 M^4 s K_{\boldsymbol{\Sigma}^*}\frac{\log d}{n} + 3\pi C_1^2 M^4 s^2 \left(\frac{\log d}{n}\right)^{3/2}$$
$$\leq M^2 K_{\boldsymbol{\Sigma}^*}\sqrt{\frac{\log d}{N}} + \left(6\pi C_1 M^3 + 4C_0 C_1 M^4 K_{\boldsymbol{\Sigma}^*} + 3\pi C_1^2 M^4 s\sqrt{\frac{\log d}{n}}\right)\frac{s\log d}{n}$$
$$= M^2 K_{\boldsymbol{\Sigma}^*}\sqrt{\frac{\log d}{N}} + C' M^4 K_{\boldsymbol{\Sigma}^*}\frac{s\log d}{n}.$$

which holds with probability at least $1 - 2/d - 1/d^{C_2-2}$, where $C', C_2$ is an absolute constant and the second inequality is due to the fact that $s\sqrt{\log d/n} = o_p(1)$ by Lemma B.2. □

## D  Auxiliary Lemmas

First, we present the results of estimation error for CLIME method.

**Lemma D.1.** Suppose $\boldsymbol{\Theta}^* \in \mathcal{U}(s, M)$ and $\lambda \geq M\|\widehat{\boldsymbol{\Sigma}} - \boldsymbol{\Sigma}^*\|_{\infty,\infty} = C_0\nu\sqrt{\log d/n}$ for some constant $C_0 > 0$, where $\widehat{\boldsymbol{\Sigma}}$ is the sample covariance matrix based on $n$ samples. Then for the CLIME estimator $\widehat{\boldsymbol{\Theta}}$ in (3.3), we have with probability at least $1 - d^{-1}$ that

$$\|\widehat{\boldsymbol{\Theta}} - \boldsymbol{\Theta}^*\|_{\infty,\infty} \leq 4C_0 M^2 \sqrt{\frac{\log d}{n}}, \qquad \|\widehat{\boldsymbol{\Theta}} - \boldsymbol{\Theta}^*\|_1 \leq C_1 M^2 s\sqrt{\frac{\log d}{n}},$$

where $C_1$ is a constant only depending on $C_0$.

Lemma D.1 is proved by Cai et al. (2011). We remark that it shows CLIME has strong guarantees in its estimation error, in terms of different matrix norms. Note that it is also able to derive estimation error bound in terms of spectral norm $\|\cdot\|_2$ and elementwise sup norm $\|\cdot\|_{\infty,\infty}$ for graphical Lasso as shown in Ravikumar et al. (2011). However, it requires irrepresentable condition, which is a very stringent assumption.

**Lemma D.2** (Jankova and van de Geer (2013)). Let $\mathbf{a}, \mathbf{b} \in \mathbb{R}^d$ such that $\|\mathbf{a}\|_2, \|\mathbf{b}\|_2 \leq M$. Let $\boldsymbol{X} \in \mathbb{R}^d$ be a $K$-subGaussian random vector. Then for $l \geq 2$,

$$\mathbb{E}\big|\mathbf{a}^\top \boldsymbol{X}\boldsymbol{X}^\top \mathbf{b} - \mathbb{E}(\mathbf{a}^\top \boldsymbol{X}\boldsymbol{X}^\top \mathbf{b})\big|^l \leq \frac{l!(2M^2 K^2)^l}{2}.$$

**Lemma D.3.** For $X_1$ and $X_2$ being two sub-Gaussian random variables, $X_1 \cdot X_2$ is a sub-exponential random variable with

$$\|X_1 \cdot X_2\|_{\psi_1} \leq C \cdot \max\big\{\|X_1\|_{\psi_2}^2, \ \|X_2\|_{\psi_2}^2\big\},$$

where $C > 0$ is an constant.

**Lemma D.4** (Bernstein-type inequality (Vershynin, 2010)). Let $X_1, \cdots, X_n$ be independent centered sub-exponential random variables, and $M = \max_{1 \leq i \leq n} \|X_i\|_{\psi_1}$. Then for every $\boldsymbol{a} = (a_1, \cdots, a_n)^\top \in \mathbb{R}^n$ and every $t \geq 0$ we have

$$\mathbb{P}\bigg(\sum_{i=1}^n a_i X_i \geq t\bigg) \leq \exp\bigg\{-C_2 \min\bigg\{\frac{t^2}{M^2\|\boldsymbol{a}\|_2^2}, \frac{t}{M\|\boldsymbol{a}\|_\infty}\bigg\}\bigg\}.$$

**Lemma D.5** (McDiarmid's inequality (Boucheron et al., 2013)). Let $Z_1, \ldots, Z_n$ be independent random variables. Suppose that

$$\sup_{z_1,\ldots,z_n,z_{i'}} |f(z_1,\ldots,z_i,\ldots,z_n) - f(z_1,\ldots,z_{i'},\ldots,z_n)| \leq c_i,$$

where $c_i$ is a constant, $i = 1, \ldots, n$. Then it holds that

$$\mathbb{P}\big(|f(Z_1,\ldots,Z_n) - \mathbb{E}\big(f(Z_1,\ldots,Z_n)\big)| > t\big) \leq 2\exp\bigg\{-\frac{2t^2}{\sum_{i=1}^n c_i^2}\bigg\},$$

for $\forall t > 0$.

**Lemma D.6** (Han and Liu (2012)). Given i.i.d. random vectors $\boldsymbol{X}_1, \boldsymbol{X}_2, \ldots, \boldsymbol{X}_n$ following $TE_d(\boldsymbol{\Sigma}^*, \xi; f_1, f_2, \ldots, f_d)$, let $\widehat{\boldsymbol{\Sigma}}$ be the Kendall's tau correlation matrix, we have

$$\big\|\widehat{\boldsymbol{\Sigma}} - \boldsymbol{\Sigma}^*\big\|_{\infty,\infty} \leq 3\pi\sqrt{\frac{\log d}{n}}$$

holds with probability at least $1 - d^{-5/2}$.